\documentclass[11pt]{article}

\usepackage[final]{acl}
\usepackage{times}
\usepackage{latexsym}
\usepackage[T1]{fontenc}
\usepackage[utf8]{inputenc}
\usepackage{microtype}
\usepackage{inconsolata}
\usepackage{graphicx}
\usepackage{subcaption}
\usepackage{booktabs} 
\usepackage{url}
\usepackage{multirow}
\newcommand{\quotes}[1]{``#1''}
\usepackage{arydshln}
\usepackage{wrapfig}
\usepackage{caption}
\usepackage{amsmath}

\usepackage[xcdraw]{xcolor}
\usepackage{algorithm}
\usepackage{algorithmic}
\usepackage{amssymb}

\newcommand\blfootnote[1]{%
  \begingroup
  \renewcommand\thefootnote{}\footnote{#1}%
  \addtocounter{footnote}{-1}%
  \endgroup
}

\title{TripPattern: A Pattern-based Text Watermarking Method \\for Large Language Models}

\author{Sangjun Moon$^{1}$, Dasom Choi$^{1}$, Jingun Kwon$^{1*}$, \\ \textbf{Hidetaka Kamigaito}$^2$, \textbf{Taro Watanabe}$^2$, \textbf{and Manabu Okumura}$^3$ \\
 $^1$Chungnam National University, $^2$Nara Institute of Science and Technology (NAIST) \\
 $^3$Institute of Science Tokyo \\
 {\tt \{sangjunmoon, dasomchoi\}@o.cnu.ac.kr} \\
 {\tt jingun.kwon@cnu.ac.kr} \\
 {\tt \{kamigaito.h, taro\}@is.naist.jp}\\
 {\tt oku@pi.titech.ac.jp}
 \\}

\begin{document}
\maketitle
\blfootnote{$^*$ Corresponding author.}
\begin{abstract}
Text watermarking techniques have gained significant attention for identifying machine-generated text and mitigating risks from large language models (LLMs). Existing methods typically divide an LLM’s vocabulary into green and red tokens, but encouraging generation toward green tokens can reduce text quality and naturalness. To address this, we propose \textbf{TripPattern}, a watermarking framework that formulates text watermarking as a pattern-based matching task using three vocabulary partitions. TripPattern divides the vocabulary into one neutral group and two pattern groups. During generation, the model alternates token selection between the two pattern groups to embed detectable patterns, while neutral tokens are selected independently to improve flexibility and preserve naturalness. For detection, TripPattern uses pattern-based statistical tests that provide interpretable p-values by measuring how often adjacent tokens alternate between the pattern groups. Theoretical analysis and empirical evaluations on four multilingual datasets show that TripPattern maintains LLM generation quality while achieving robust watermark detectability.
\end{abstract}

\section{Introduction}
Large language models (LLMs) have demonstrated strong generative capabilities across diverse domains~\cite{10.5555/3600270.3602281,openai2024gpt4technicalreport,grattafiori2024llama3herdmodels,qwen2025qwen25technicalreport}, but their widespread use has raised concerns about misinformation~\cite{10.5555/3454287.3455099,chen-etal-2023-beyond}, academic dishonesty~\cite{StokelWalker2022AIBC,vasilatos2025howkgptinvestigatingdetectionchatgptgenerated}, and the reliability of digital content~\cite{tang-etal-2024-tofueval,li-etal-2024-dawn}. These concerns highlight the need for robust watermarking methods that can reliably identify machine-generated text, especially as LLM-generated content may affect future models trained on artificial data~\cite{radford2022robustspeechrecognitionlargescale,shumailov2024curserecursiontraininggenerated}.

\begin{figure}[t!]
\centering
    \includegraphics[width=0.85\columnwidth]{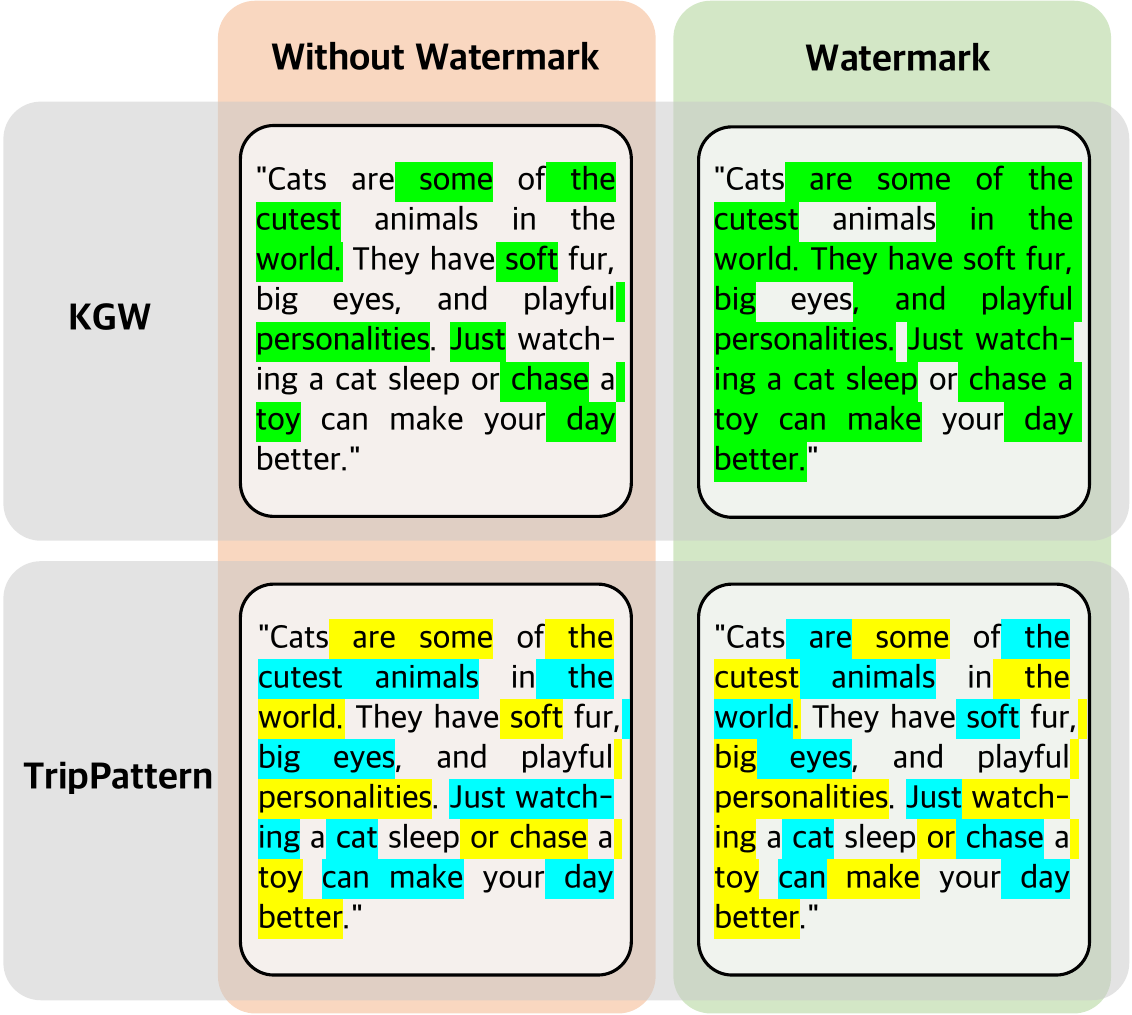}
    \caption{Unlike KGW’s two-group partition, which encourages green token generation, TripPattern alternates between two pattern groups while allowing neutral-token sampling to preserve text quality.}
    \label{fig:intro}
\end{figure}

Decoding-based watermarking methods embed identifiable signals directly into generated text and have received significant attention for detecting LLM-generated content~\cite{sadasivan2025aigeneratedtextreliablydetected}. Most existing approaches rely on the KGW-style two-group token partitioning framework, which divides the vocabulary into green and red tokens~\cite{kirchenbauer2024watermarklargelanguagemodels,guo-etal-2024-context-aware,ren-etal-2024-robust,chen-etal-2024-watme,zhao2024provable,wu-chandrasekaran-2024-bypassing}. By biasing generation toward green tokens, these methods encode detectable fingerprints; however, this restriction reduces vocabulary flexibility and often degrades text quality and naturalness. This limitation is particularly problematic in multilingual and practical settings, where linguistic diversity and fluency are important.

Although several studies have attempted to reduce quality degradation through unbiased distribution adjustments~\cite{kuditipudi2024robust,hu2024unbiased}, entropy-based gating~\cite{lu-etal-2024-entropy,10.5555/3692070.3693308,lee-etal-2024-wrote}, dynamic watermark-strength scaling~\cite{wang-etal-2025-morphmark}, or auxiliary models~\cite{liu2024an,he-etal-2024-watermarks,10.5555/3692070.3692903}, these methods often introduce trade-offs among text quality, detection robustness, computational cost, and deployment flexibility~\cite{chen-etal-2024-watme,mao-etal-2025-watermarking}.

To address these limitations, we propose \textbf{TripPattern}, a watermarking approach that formulates watermark embedding as a pattern-based matching task within a three-group vocabulary partitioning framework. As shown in Figure~\ref{fig:intro}, TripPattern divides the vocabulary into one neutral group and two distinct pattern groups. During generation, the model alternates token selection between the two pattern groups to embed statistically detectable patterns, while tokens from the neutral group can be sampled independently to preserve fluency and naturalness.

To construct the neutral group, we further consider subword token length and assign tokens shorter than a predefined threshold to this group. Since shorter tokens often occur more frequently due to linguistic economy~\cite{KANWAL201745}, and tokenization algorithms tend to encode frequent morphemes or words as shorter subword units~\cite{chung-etal-2020-improving}, this design helps maintain grammatical coherence and can be naturally applied to multilingual scenarios.

For detection, TripPattern performs a pattern-based statistical test using only tokens from the two pattern groups, ignoring neutral tokens that do not carry watermark signals. The detector measures how often adjacent pattern-group tokens alternate and uses this statistic to determine whether the text contains the embedded watermark.

Through theoretical analysis and experiments on four multilingual datasets, we show that TripPattern preserves the generative quality of LLMs while achieving robust watermark detectability.

\section{Method}
In this section, we first review the widely used two-group watermarking framework. We then present \textbf{TripPattern}, which employs a three-group vocabulary partition and constructs a neutral group based on subword token length.

\subsection{Preliminary}
Text watermarking for LLMs generally consists of two procedures: watermark encoding and detection. Given an input prompt $\mathbf{x}_{1:n} = \{x_1, x_2, ..., x_n\}$, an LLM $\mathcal{M}$ generates an output sequence $\{x_{n+1}, x_{n+2}, ..., x_m\}$, into which the encoding procedure embeds an identifiable watermark.

A representative approach is the KGW-based two-group partitioning framework~\cite{kirchenbauer2024watermarklargelanguagemodels}. At each generation step $t$, a keyed hash function computes
\[
h_{n+t}=H_k(\mathbf{x}_{n+t-1}),
\]
where $H_k$ is a cryptographic hash or PRF keyed by a secret key $k$. The resulting hash value seeds a pseudorandom number generator (PRNG), which deterministically partitions the vocabulary $\mathcal{V}$ into two disjoint subsets: green tokens $\mathcal{V}_g$ and red tokens $\mathcal{V}_r$.

Let \( l^{n+t} \in \mathbb{R}^{|\mathcal{V}|} \) denote the logits predicted by $\mathcal{M}$ at step $t$. The KGW framework adds a positive bias $\delta$ to green-token logits:
\begin{equation}
l_i^{n+t} =
\begin{cases}
l_i^{n+t} + \delta, & \text{if } v_i \in \mathcal{V}_g,\\[6pt]
l_i^{n+t}, & \text{if } v_i \in \mathcal{V}_r,
\end{cases}
\end{equation}
where \(i\) denotes the token index in the vocabulary. As a result, watermarked text statistically contains more green tokens than ordinary human-written text.

For detection, the verifier recomputes the same partitioning process using the shared hash function and PRNG, and calculates the following z-score:
\begin{equation}
 z=(G-\gamma L)/\sqrt{L\gamma(1-\gamma)},   
\end{equation}
where $G$ is the number of green tokens in the generated output, $L$ is the total number of generated tokens, and $\gamma$ is the green-token ratio. The watermark is identified by testing the null hypothesis \textit{$H_0$: The generated text sequence does not follow the green-list constraint}. If the computed z-score exceeds a predefined threshold, $H_0$ is rejected.

Despite its effectiveness, the two-group framework inherently reduces token-selection flexibility by encouraging the model to select green tokens, which can degrade text quality and naturalness. Several studies have attempted to mitigate this issue through unbiased output-distribution adjustments, such as Gumbel softmax and output reweighting~\cite{kuditipudi2024robust,hu2024unbiased}. However, preserving an unbiased distribution does not necessarily guarantee both text quality and detection performance, especially for aligned LLMs~\cite{chen-etal-2024-watme}.

\subsection{TripPattern: Three-group Partitioning}
We propose \textbf{TripPattern}, a watermarking method that formulates watermark embedding as a pattern-based matching task. Unlike the conventional green/red partition, TripPattern divides the vocabulary into three disjoint subsets: one neutral group $\mathcal{V}_n$ and two pattern groups $\mathcal{V}_{p1}$ and $\mathcal{V}_{p2}$. The neutral group contains freely selectable tokens, while the two pattern groups are used to embed detectable alternation patterns.

Following prior watermarking methods~\cite{kuditipudi2024robust}, TripPattern uses a keyed hash and PRNG to deterministically partition the vocabulary at each generation step. The same procedure is used during both generation and detection, ensuring that the partition can be reconstructed by an authorized detector.

The encoding process encourages an alternating pattern between the two pattern groups. Specifically, if the last non-neutral token belongs to $\mathcal{V}_{p1}$, the next allowable set is biased toward $\mathcal{V}_{p2} \cup \mathcal{V}_n$; conversely, if the last non-neutral token belongs to $\mathcal{V}_{p2}$, the next allowable set is biased toward $\mathcal{V}_{p1} \cup \mathcal{V}_n$. Neutral tokens do not change the current alternation state, allowing the model to preserve fluency while maintaining the watermark pattern.

At each generation step, TripPattern applies a positive soft bias $\delta>0$ to the logits of tokens in the allowable set $\mathcal{A}_t$ and then normalizes over the full vocabulary $\mathcal{V}$. This design statistically encourages the intended alternation pattern while retaining broader token-selection flexibility than two-group partitioning. 

\subsection{Subword Length-based Neutral Group Construction}
We further construct the neutral group using subword token length at each decoding step.
Given the step-specific hash value, the PRNG deterministically repartitions the vocabulary into
$\mathcal{V}_n$, $\mathcal{V}_{p1}$, and $\mathcal{V}_{p2}$.
During this per-step partitioning process, tokens shorter than a predefined threshold are prioritized
for assignment to $\mathcal{V}_n$, and the remaining neutral slots are filled according to the PRNG
until the target neutral-group ratio $\gamma$ is reached.
The remaining tokens are then divided into the two pattern groups.

Thus, the neutral group is not a fixed global vocabulary subset; rather, it is reconstructed at each
generation and detection step using the same keyed hash/PRNG procedure, with subword length serving
as a deterministic priority rule. This ensures that the detector can reproduce the same step-specific
partitions while allowing frequent short subword tokens to more often function as neutral tokens.

\subsection{Detecting the Watermark}
TripPattern detects watermarks by testing whether a given text follows the alternation pattern induced during generation. Given a text sequence, the detector reconstructs the three-way partition using the same keyed hash and PRNG. It then removes tokens belonging to the neutral group and maps the remaining tokens to their corresponding pattern labels, such as \quotes{1} and \quotes{2}.

The detector counts the number of adjacent pattern-token pairs and the number of alternations between $\mathcal{V}_{p1}$ and $\mathcal{V}_{p2}$. Let $C$ denote the number of observed alternations and $r$ the corresponding alternation ratio. The null hypothesis is defined as \textit{$H_0$: The generated text sequence does not follow the alternation pattern}. The observed statistic is tested under the permutation null conditional on $(n_1,n_2)$ using the runs moments~\cite{wald1940runs}. If the resulting z-score exceeds a predefined threshold, $H_0$ is rejected and the watermark is identified. The detailed generation and detection algorithms are described in Appendix~\ref{appen:alg}.

\subsection{Theoretical Analysis}
We provide a theoretical analysis showing that TripPattern offers greater token-selection flexibility than the KGW-based two-group framework.

\noindent \textbf{Definition (Vocabulary Partition).}
\textit{
Let $\mathcal{V}$ denote the vocabulary of the language model. In the KGW-based framework, the vocabulary is partitioned into two disjoint subsets:
\[
\mathcal{V}=\mathcal{V}_g \cup \mathcal{V}_r,
\qquad
\mathcal{V}_g \cap \mathcal{V}_r = \emptyset,
\]
where $\mathcal{V}_g$ and $\mathcal{V}_r$ denote the green and red token sets, respectively. In TripPattern, the vocabulary is partitioned into three disjoint subsets:
\[
\mathcal{V}=\mathcal{V}_{p1} \cup \mathcal{V}_{p2} \cup \mathcal{V}_n,
\]
where $\mathcal{V}_{p1}$ and $\mathcal{V}_{p2}$ are the two pattern groups and $\mathcal{V}_n$ is the neutral group.
}

\noindent \textbf{Theorem 1. (Greater selection flexibility of TripPattern)}
\textit{
Let $\mathcal{A}^{KGW}_t$ and $\mathcal{A}^{Trip}_t$ denote the biased token sets at decoding step $t$ under KGW and TripPattern, respectively. Let $\gamma_g$ be the green-token ratio in KGW and $\gamma_n$ be the neutral-group ratio in TripPattern. Then,
\[
|\mathcal{A}^{KGW}_t| = \gamma_g |\mathcal{V}|,
\qquad
|\mathcal{A}^{Trip}_t| = \frac{1+\gamma_n}{2}|\mathcal{V}|.
\]
Therefore,
\[
|\mathcal{A}^{Trip}_t| \ge |\mathcal{A}^{KGW}_t|
\quad \text{if} \quad
\frac{1+\gamma_n}{2} \ge \gamma_g,
\]
with strict inequality when $\frac{1+\gamma_n}{2} > \gamma_g$.
See Appendix~\ref{appen:proof} for the proof.
}

Thus, TripPattern provides a larger biased token set than KGW under this condition, suggesting greater token-selection flexibility. We empirically examine whether this flexibility translates into improved generation quality.

\begin{table*}[t!]
\renewcommand{\arraystretch}{0.8}
\centering

\resizebox{0.99\textwidth}{!}{
\begin{tabular}{lccccccccccccccc}
\toprule
 & \multicolumn{3}{c}{\textbf{English}} & \multicolumn{3}{c}{\textbf{Korean}}  & \multicolumn{3}{c}{\textbf{German}} & \multicolumn{3}{c}{\textbf{Spanish}} & \multicolumn{3}{c}{\textbf{Average}} \\
\cmidrule(lr){2-4} \cmidrule(lr){5-7} \cmidrule(lr){8-10} \cmidrule(lr){11-13} \cmidrule(lr){14-16}
\multirow{-2}{*}{\textbf{Model}} & \textbf{PPL $\downarrow$} & \textbf{AUC $\uparrow$} & \textbf{TPR@5 $\uparrow$} & \textbf{PPL $\downarrow$} & \textbf{AUC $\uparrow$} & \textbf{TPR@5 $\uparrow$} & \textbf{PPL $\downarrow$} & \textbf{AUC $\uparrow$} & \textbf{TPR@5 $\uparrow$} & \textbf{PPL $\downarrow$} & \textbf{AUC $\uparrow$} & \textbf{TPR@5 $\uparrow$} & \textbf{PPL $\downarrow$} & \textbf{AUC $\uparrow$} & \textbf{TPR@5 $\uparrow$} \\
\midrule
Llama-3.2-3B  & 3.49 & - & - & 3.86 & - & - & 3.53 & - & - & 3.14 & - & - & 3.51 & - & - \\
\cline{2-16}
\addlinespace[0.2em] 
 +KGW     & 5.83 & 0.996 & 0.990 & 7.48 & 0.987 & 0.966 & 6.41 & 0.990 & 0.972 & 5.48 & 0.993 & 0.984 & 6.30 & 0.991 & 0.978 \\
 +Unigram & 6.55 & 0.998 & 0.996 & \textbf{4.86} & 0.988 & 0.962 & 6.07 & 0.993 & \textbf{0.984} & 5.11 & 0.992 & 0.976 & 5.65 & 0.993 & 0.980 \\
 +UPV     & 5.36 & \textbf{1.000} & 0.998 & 5.70 & 0.960 & 0.858 & \textbf{5.65} & 0.993 & 0.978 & 4.95 & 0.986 & 0.972 & \textbf{5.42} & 0.985 & 0.952 \\
 +SWEET   & 5.25 & 0.996 & 0.990 & 7.07 & 0.989 & 0.966 & 5.68 & 0.990 & 0.970 & 5.07 & 0.993 & 0.982 & 5.77 & 0.992 & 0.977 \\
 \addlinespace[0.1em]            
 \cdashline{2-16}
 \addlinespace[0.3em]
 +TripPattern         & \textbf{5.24} & \textbf{1.000} & \textbf{1.000} & 6.33 & \textbf{0.995} & 0.968 & \textbf{5.65} & \textbf{0.996} & 0.982 & \textbf{4.53} & 0.988 & 0.968 & 5.44 & \textbf{0.995} & 0.980 \\
 +TripPattern w/o Len & 5.26 & 0.998 & 0.990 & 7.88 & 0.986 & \textbf{0.974} & 6.00 & 0.990 & 0.972 & 5.18 & \textbf{0.996} & \textbf{0.988} & 6.08 & 0.993 & \textbf{0.981}\\
\midrule
Llama-3.1-8B  & 3.08 & - & - & 3.06 & - & - & 3.07 & - & - & 2.66 & - & - & 2.97 & - & - \\
\cline{2-16}
\addlinespace[0.2em] 
 +KGW      & 5.16 & \textbf{1.000} & \textbf{1.000} & 6.03 & 0.998 & \textbf{0.996} & 5.24 & 0.998 & 0.996 & 4.29 & 0.998 & 0.998 & 5.18 & 0.999 & \textbf{0.998} \\
 +Unigram  & 5.58 & \textbf{1.000} & \textbf{1.000} & 5.16 & 0.996 & 0.988 & 5.00 & 0.996 & 0.992 & 4.29 & 0.997 & 0.996 & 5.01 & 0.997 & 0.994 \\
 +UPV      & 4.66 & 0.999 & 0.994 & \textbf{4.55} & 0.962 & 0.926 & 4.72 & 0.994 & 0.984 & 3.95 & 0.997 & 0.984 & 4.47 & 0.988 & 0.972 \\
 +SWEET    & \textbf{4.61} & 0.999 & 0.998 & 5.69 & 0.998 & 0.988 & 4.78 & 0.998 & 0.996 & 3.94 & \textbf{1.000} & \textbf{1.000} & 4.76 & 0.999 & 0.996 \\
 \addlinespace[0.1em]            
 \cdashline{2-16}
 \addlinespace[0.3em]
 +TripPattern         & \textbf{4.61} & \textbf{1.000} & \textbf{1.000} & 4.63 & 0.990 & 0.970 & \textbf{4.45} & 0.993 & 0.986 & \textbf{3.63} & 0.995 & 0.992 & \textbf{4.33} & 0.995 & 0.987 \\
 +TripPattern w/o Len & \textbf{4.61} & 0.999 & 0.996 & 5.23 & \textbf{0.999} & \textbf{0.996} & 4.80 & \textbf{1.000} & \textbf{0.999} & 3.88 & \textbf{1.000} & 0.999 & 4.63 & \textbf{1.000} & \textbf{0.998} \\
\midrule
Qwen-2.5-3B  & 3.93 & - & - & 3.30 & - & - & 3.68 & - & - & 3.19 & - & - & 3.53 & - & - \\
\cline{2-16}
\addlinespace[0.2em] 
 +KGW      & 8.96 & 0.992 & 0.982 & 7.22 & 0.968 & 0.930 & 8.08 & 0.966 & 0.932 & 6.13 & 0.963 & 0.924 & 7.60 & 0.972 & 0.942 \\
 +Unigram  & 8.03 & 0.990 & 0.964 & 6.55 & 0.955 & 0.892 & 6.45 & 0.951 & 0.878 & 5.32 & 0.927 & 0.808 & 6.59 & 0.956 & 0.886 \\
 +UPV      & 7.82 & 0.991 & 0.978 & 5.56 & 0.943 & 0.874 & 6.40 & 0.952 & 0.898 & 5.22 & 0.945 & 0.856 & 6.25 & 0.958 & 0.902 \\
 +SWEET    & 8.28 & 0.988 & 0.974 & 6.27 & 0.949 & 0.904 & 7.33 & 0.954 & 0.910 & 5.52 & 0.940 & 0.874 & 6.85 & 0.958 & 0.916 \\
 \addlinespace[0.1em]            
 \cdashline{2-16}
 \addlinespace[0.3em]
 +TripPattern         & 7.05 & \textbf{0.997} & \textbf{0.994} & \textbf{4.74} & \textbf{0.985} & 0.956 & \textbf{6.30} & 0.986 & 0.964 & \textbf{4.91} & \textbf{0.969} & 0.924 & \textbf{5.75} & \textbf{0.984} & 0.960\\
 +TripPattern w/o Len & \textbf{6.80} & 0.994 & 0.984 & 6.17 & 0.984 & \textbf{0.962} & 7.04 & \textbf{0.989} & \textbf{0.966} & 5.37 & \textbf{0.969} & \textbf{0.940} & 6.35 & \textbf{0.984} & \textbf{0.963}\\
\midrule
Qwen-2.5-7B  & 3.24 & - & - & 3.17 & - & - & 3.49 & - & - & 2.90 & - & - & 3.20 & - & - \\
\cline{2-16}
\addlinespace[0.2em] 
 +KGW     & 5.73 & \textbf{1.000} & \textbf{1.000} & 5.86 & 0.992 & 0.986 & 6.49 & 0.993 & \textbf{0.992} & 4.70 & 0.992 & 0.986 & 5.70 & 0.994 & 0.991\\
 +Unigram & 5.50 & 0.997 & 0.994 & 5.55 & 0.991 & \textbf{0.990} & 6.39 & 0.991 & 0.984 & 4.86 & 0.989 & 0.984 & 5.58 & 0.992 & 0.988\\
 +UPV     & 5.25 & 0.999 & 0.994 & 4.66 & 0.980 & 0.948 & 5.39 & 0.982 & 0.970 & 4.33 & 0.987 & 0.982 & 4.91 & 0.987 & 0.974\\
 +SWEET   & 5.27 & 0.996 & 0.994 & 5.32 & 0.995 & 0.988 & 5.90 & 0.992 & 0.982 & \textbf{4.26} & 0.991 & 0.978 & 5.19 & 0.993 & 0.986\\
 \addlinespace[0.1em]            
 \cdashline{2-16}
 \addlinespace[0.3em]
 +TripPattern         & 5.33 & 0.999 & 0.996 & 5.02 & 0.992 & 0.986 & \textbf{5.35} & 0.992 & 0.986 & \textbf{4.26} & 0.995 & 0.988 & \textbf{4.80} & 0.995 & 0.989 \\
 +TripPattern w/o Len & \textbf{5.07} & \textbf{1.000} & \textbf{1.000} & \textbf{4.24} & \textbf{0.997} & 0.986 & 5.54 & \textbf{0.997} & \textbf{0.992} & \textbf{4.26} & \textbf{0.996} & \textbf{0.996} & 4.97 & \textbf{0.998} & \textbf{0.994}\\
\bottomrule
\end{tabular}}
\caption{Experimental results comparing TripPattern with baseline watermarking methods. TripPattern w/o Len is a method that does not consider subword length when defining the neutral group.
Lower PPL scores indicate better text quality, whereas higher AUC and TPR@5 scores indicate better detection performance.}
\label{tab:c4_multilingual_full_results}

\end{table*}

\section{Experiments}
\subsection{Experimental Settings}
\noindent \textbf{Datasets.}
We evaluate TripPattern across four languages: English, Korean, German, and Spanish. For English fluency evaluation, we use the C4 dataset~\cite{dodge-etal-2021-documenting}, which is widely used in text watermarking studies. For multilingual fluency evaluation, we use mC4, a multilingual web-crawled corpus from Common Crawl~\cite{xue-etal-2021-mt5}. Following prior work, we randomly sample 500 instances from each test dataset~\cite{kirchenbauer2024watermarklargelanguagemodels,guo-etal-2024-context-aware,kuditipudi2024robust}.

For mathematical reasoning evaluation, we use GSM8K~\cite{cobbe2021trainingverifierssolvemath}, which contains 1,319 test instances of grade-school math problems. We follow the CoT Hub prompting protocol~\cite{fu2023chainofthoughthubcontinuouseffort} and use few-shot prompts that require models to generate reasoning steps before producing final answers. For multilingual reasoning, we use MGSM~\cite{shi2022languagemodelsmultilingualchainofthought}, which contains 250 test instances per language. Following the MGSM setup, we use eight training instances as few-shot examples. Since Korean is not included in the original MGSM benchmark, we use a translated version of GSM8K.\footnote{\url{https://huggingface.co/datasets/ChuGyouk/GSM8k-Ko}}

\noindent \textbf{Evaluation Metrics.}
Following prior work~\cite{ren-etal-2024-robust,kirchenbauer2024watermarklargelanguagemodels,guo-etal-2024-context-aware,chen-etal-2024-watme,lu-etal-2024-entropy}, we evaluate both generation quality and watermark detectability. For fluency evaluation, we report perplexity (PPL) on C4. For mathematical reasoning, we measure accuracy based on the correctness of generated answers. For watermark detection, we report the Area Under the Receiver Operating Characteristic curve (AUC) and true positive rate at a false positive rate of at most 5\% (TPR@5).

\noindent \textbf{Baselines.}
We compare \textbf{TripPattern} with several state-of-the-art watermarking baselines. \textbf{TripPattern w/o Len} is an ablated variant that removes subword length-based neutral group construction. \textbf{KGW} is a standard watermarking method that partitions the vocabulary into green and red token groups~\cite{kirchenbauer2024watermarklargelanguagemodels}. \textbf{Unigram} uses a fixed green list determined by a hash key to improve robustness~\cite{zhao2024provable}. \textbf{UPV} extends KGW with a cryptographically secure public verification scheme using an auxiliary detector~\cite{liu2024an}. \textbf{SWEET} applies watermarking only in high-entropy contexts~\cite{lee-etal-2024-wrote}. We also evaluate \textbf{Morphmark}, which adaptively scales the watermark bias according to token-level green probability mass~\cite{wang-etal-2025-morphmark}; its results are reported in Appendix~\ref{appen:morphmark} due to limited detection performance.

\noindent \textbf{Models and Implementation Details.}
We conduct experiments using instruction-tuned models from the Llama-3 family~\cite{grattafiori2024llama3herdmodels} with 3B and 8B parameters, and the Qwen-2.5 family~\cite{qwen2025qwen25technicalreport} with 3B and 7B parameters. For PPL evaluation, we use Qwen-2.5-14B-Instruct as the evaluator. Following prior work~\cite{kirchenbauer2024watermarklargelanguagemodels,chen-etal-2024-watme}, we use greedy decoding in all experiments to remove sampling randomness.
We set the subword length threshold for the neutral group to three across all languages. For KGW-based methods, we set $\delta=3.0$ and $\gamma=0.3$; for pattern-based methods, we set $\delta=5.0$ and $\gamma=0.3$. The minimum and maximum generation lengths are set to 50 and 200 tokens, respectively. For KGW-based methods, $\gamma$ denotes the green-token ratio, whereas for TripPattern, $\gamma$ denotes the neutral-group ratio.

\begin{table*}[t!]
\renewcommand{\arraystretch}{0.8}
\centering

\resizebox{0.99\textwidth}{!}{
\begin{tabular}{lccccccccccccccc}
\toprule
 & \multicolumn{3}{c}{\textbf{English}} & \multicolumn{3}{c}{\textbf{Korean}} & \multicolumn{3}{c}{\textbf{German}} & \multicolumn{3}{c}{\textbf{Spanish}} & \multicolumn{3}{c}{\textbf{Average}} \\
\cmidrule(lr){2-4} \cmidrule(lr){5-7} \cmidrule(lr){8-10} \cmidrule(lr){11-13} \cmidrule(lr){14-16}
 \multirow{-2}{*}{\textbf{Model}} & \textbf{ACC $\uparrow$} & \textbf{AUC $\uparrow$} & \textbf{TPR@5 $\uparrow$} & \textbf{ACC $\uparrow$} & \textbf{AUC $\uparrow$} & \textbf{TPR@5 $\uparrow$} & \textbf{ACC $\uparrow$} & \textbf{AUC $\uparrow$} & \textbf{TPR@5 $\uparrow$} & \textbf{ACC $\uparrow$} & \textbf{AUC $\uparrow$} & \textbf{TPR@5 $\uparrow$} & \textbf{ACC $\uparrow$} & \textbf{AUC $\uparrow$} & \textbf{TPR@5 $\uparrow$}\\
\midrule

Llama-3.2-3B  & 72.4 & - & - & 7.35 & - & - & 11.6 & - & - & 11.45 & - & - & 25.70 & - & - \\
\cline{2-16}
\addlinespace[0.2em] 
 +KGW                 & 64.06 & 0.963 & 0.814 & 6.29 & 0.982 & 0.900 & 9.78  & 0.942 & 0.700 & 11.30 & 0.926 & 0.692 & 22.86 & 0.953 & 0.777 \\
 +Unigram             & 61.64 & 0.938 & 0.724 & 6.44 & 0.960 & 0.828 & 9.70  & 0.949 & 0.780 & 10.46 & 0.906 & 0.648 & 22.06 & 0.938 & 0.745 \\
 +UPV                 & 61.41 & 0.927 & 0.702 & 5.61 & 0.932 & 0.748 & 8.64  & 0.943 & 0.760 & 10.39 & 0.894 & 0.620 & 21.51 & 0.924 & 0.708 \\
 +SWEET               & \textbf{67.75} & 0.920 & 0.697 & 6.67 & \textbf{0.989} & \textbf{0.952} & 10.31 & 0.933 & 0.748 & \textbf{11.52} & 0.903 & 0.676 & 24.06 & 0.936 & 0.768\\
 \addlinespace[0.1em]            
 \cdashline{2-16}
 \addlinespace[0.3em]
 +TripPattern         & 67.10 & 0.950 & 0.776 & \textbf{7.96} & 0.965 & 0.848 & \textbf{10.54} & \textbf{0.966} & \textbf{0.816} & 11.45 & 0.932 & 0.648 & \textbf{24.26} & 0.953 & 0.772\\
 +TripPattern w/o Len & 64.22 & \textbf{0.964} & \textbf{0.817} & 5.46 & 0.977 & 0.892 & 9.78  & 0.946 & 0.784 & 9.78  & \textbf{0.938} & \textbf{0.760} & 22.31 & \textbf{0.956} & \textbf{0.813} \\
\midrule
Llama-3.1-8B  & 81.05 & - & - & 10.31 & - & - & 13.65 & - & - & 13.95 & - & - & 29.74 & - & - \\
\cline{2-16}
\addlinespace[0.2em] 
 +KGW                 & 68.76 & 0.960 & 0.826 & 8.79 & \textbf{0.984} & \textbf{0.924} & 12.13 & 0.972 & 0.796 & 11.98 & 0.948 & 0.748 & 25.42 & \textbf{0.966} & 0.824\\
 +Unigram             & 68.99 & 0.940 & 0.734 & 7.81 & 0.956 & 0.820 & 11.14 & 0.972 & 0.844 & 11.14 & 0.915 & 0.636 & 24.77 & 0.946 & 0.759\\
 +UPV                 & 71.19 & 0.918 & 0.676 & 8.57 & 0.932 & 0.692 & 10.92 & 0.938 & 0.756 & 12.13 & 0.907 & 0.656 & 25.70 & 0.923 & 0.695\\
 +SWEET               & \textbf{76.72} & 0.916 & 0.693 & 9.33 & 0.982 & 0.916 & 12.21 & \textbf{0.978} & \textbf{0.900} & \textbf{12.81} & 0.948 & \textbf{0.796} & 27.77 & 0.956 & \textbf{0.826}\\
 \addlinespace[0.1em]            
 \cdashline{2-16}
 \addlinespace[0.3em]
 +TripPattern         & 75.74 & 0.944 & 0.740 & \textbf{10.46} & 0.951 & 0.776 & \textbf{12.59} & 0.973 & 0.892 & 12.51 & 0.945 & 0.760 & \textbf{27.83} & 0.953 & 0.792\\
 +TripPattern w/o Len & 69.90 & \textbf{0.966} & \textbf{0.835} & 8.11  & 0.970 & 0.840 & 10.92 & 0.959 & 0.848 & 11.83 & \textbf{0.953} & 0.744 & 25.19 & 0.962 & 0.817\\
\midrule
Qwen-2.5-3B  & 33.06 & - & - & 9.25 & - & - & 6.52 & - & - & 7.28 & - & - & 14.03 & - & -\\
\cline{2-16}
\addlinespace[0.2em] 
 +KGW                 & 32.37 & 0.909 & 0.649 & 6.29 & 0.977 & 0.864 & 6.14 & 0.965 & 0.852 & 7.58 & 0.951 & 0.832 & 13.10 & 0.951 & 0.799\\
 +Unigram             & 29.11 & 0.908 & 0.611 & 6.90 & 0.956 & 0.720 & 6.14 & 0.982 & 0.916 & 6.97 & 0.964 & 0.848 & 12.28 & 0.953 & 0.774\\
 +UPV                 & 29.11 & 0.856 & 0.477 & 5.91 & 0.958 & 0.756 & 5.16 & 0.959 & 0.836 & 6.67 & 0.923 & 0.588 & 11.71 & 0.924 & 0.664\\
 +SWEET               & \textbf{34.57} & 0.898 & 0.590 & 6.82 & \textbf{0.994} & \textbf{0.972} & \textbf{6.44} & 0.979 & 0.896 & \textbf{8.19} & \textbf{0.985} & \textbf{0.932} & \textbf{14.01} & 0.964 & 0.848\\
 \addlinespace[0.1em]            
 \cdashline{2-16}
 \addlinespace[0.3em]
 +TripPattern         & 31.54 & 0.948 & \textbf{0.768} & \textbf{7.81} & 0.980 & 0.904 & 5.46 & \textbf{0.992} & \textbf{0.956} & 7.43 & 0.982 & \textbf{0.932} & 13.06 & \textbf{0.976} & \textbf{0.890}\\
 +TripPattern w/o Len & 30.93 & \textbf{0.950} & 0.755 & 6.67 & 0.988 & 0.928 & 6.29 & 0.971 & 0.880 & 6.82 & 0.977 & 0.868 & 12.68 & 0.972 & 0.858\\
\midrule
Qwen-2.5-7B  & 55.27 & - & - & 7.51 & - & - & 6.29 & - & - & 6.37 & - & - & 18.86 & - & - \\
\cline{2-16}
\addlinespace[0.2em] 
 +KGW                 & 47.08 & 0.848 & 0.434 & 7.13 & 0.932 & 0.748 & 5.46 & 0.906 & 0.576 & 6.37 & \textbf{0.893} & \textbf{0.612} & 16.51 & \textbf{0.895} & 0.593\\
 +Unigram             & 41.32 & 0.808 & 0.345 & 6.52 & 0.873 & 0.520 & \textbf{7.20} & 0.869 & 0.576 & 6.22 & 0.814 & 0.380 & 15.32 & 0.841 & 0.455\\
 +UPV                 & 47.23 & 0.822 & 0.368 & 6.52 & 0.828 & 0.388 & 6.29 & 0.822 & 0.468 & 6.07 & 0.803 & 0.316 & 16.53 & 0.818 & 0.385\\
 +SWEET               & 50.95 & 0.754 & 0.321 & \textbf{7.51} & \textbf{0.938} & \textbf{0.784} & 5.53 & 0.897 & \textbf{0.656} & \textbf{6.82} & 0.876 & 0.568 & \textbf{17.70} & 0.866 & 0.582\\
 \addlinespace[0.1em]            
 \cdashline{2-16}
 \addlinespace[0.3em]
 +TripPattern         & \textbf{52.24} & \textbf{0.870} & \textbf{0.503} & 6.44 & 0.924 & 0.716 & 5.99 & \textbf{0.911} & 0.624 & 5.99 & 0.864 & 0.536 & 17.67 & 0.892 & \textbf{0.595}\\
 +TripPattern w/o Len & 41.39 & 0.849 & 0.461 & 6.44 & 0.935 & 0.712 & 6.52 & 0.890 & 0.560 & 5.99 & 0.869 & 0.464 & 15.09 & 0.886 & 0.549\\
\bottomrule
\end{tabular}}
\caption{Experimental results comparing TripPattern and baseline watermarking methods. Higher ACC scores indicate a greater proportion of correct solutions.}
\label{tab:gsm_multilingual_full_results}

\end{table*}
 
\subsection{Main Results}
\noindent \textbf{TripPattern Preserves Text Quality Across Languages.}
Table~\ref{tab:c4_multilingual_full_results} reports the results on C4 for English and mC4 for Korean, German, and Spanish. TripPattern generally maintains low perplexity across languages and models, showing limited degradation compared to non-watermarked generation. On average across the four languages, TripPattern achieves competitive or better text quality than the baseline watermarking methods.

The improvement is particularly clear when using the subword length-based neutral group. Compared with TripPattern w/o Len, TripPattern generally improves average fluency by assigning frequent short subword tokens, such as functional words and morphemes, to the neutral group, although the effect varies across models and languages. These results suggest that introducing a neutral group enables pattern-based watermarking while reducing the quality loss typically caused by conventional two-group green/red partitioning.

\noindent \textbf{TripPattern Maintains Performance on CoT-Prompted Reasoning Tasks.}
Table~\ref{tab:gsm_multilingual_full_results} presents the results on GSM8K and MGSM. TripPattern preserves mathematical reasoning performance across languages, with limited accuracy degradation compared to baseline watermarking methods. This indicates that TripPattern can embed detectable watermarks while maintaining the model's ability to generate coherent reasoning steps and final answers.

The benefit of subword length-based neutral group construction is also observed in reasoning tasks. Across open-ended generation and reasoning tasks, TripPattern generally benefits from subword length-based neutral group construction, improving average fluency on mC4 and maintaining stronger reasoning accuracy on MGSM.

\noindent \textbf{TripPattern Achieves Competitive Detection Performance.}
Although watermarking generally involves a trade-off between text quality and detectability, TripPattern achieves AUC scores comparable to KGW while often improving text quality, as shown in Table~\ref{tab:c4_multilingual_full_results}. KGW can still yield higher TPR@5 in some settings. Moreover, on GSM8K and MGSM, TripPattern achieves competitive detection performance while better preserving reasoning accuracy in several settings, as shown in Table~\ref{tab:gsm_multilingual_full_results}.

TripPattern w/o Len slightly outperforms TripPattern in detection because the length-based neutral group assigns frequent functional tokens to the neutral group, which can dilute the watermark signal. Nevertheless, TripPattern provides a stronger overall balance between text quality and watermark detectability, demonstrating the effectiveness of pattern-based detection with a three-group vocabulary partition.

\section{Analysis}

\noindent \textbf{Flexibility of TripPattern.}
To further analyze the token-selection flexibility of TripPattern, we evaluate a \quotes{hard} partitioning setting, which is known to reduce generation quality, particularly in low-entropy contexts~\cite{kirchenbauer2024watermarklargelanguagemodels}. Unlike the soft-bias setting in Algorithm~\ref{alg:TripPattern}, this setting restricts token selection to the designated subset. For KGW, generation is restricted to the green group, while for TripPattern, generation alternates between the target pattern group and the neutral group.

\begin{figure}[t!]
\centering
    \includegraphics[width=0.45\textwidth]{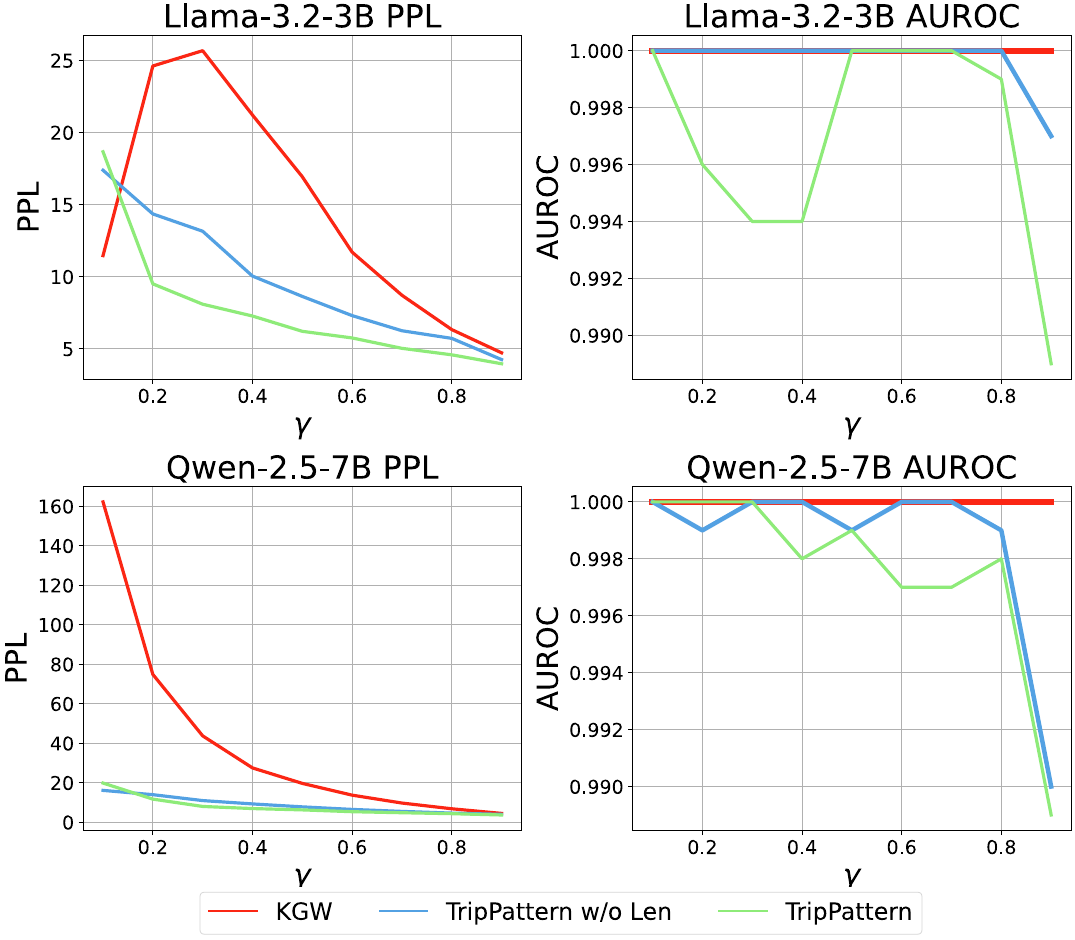}
    \caption{Comparison of TripPattern, TripPattern w/o Len, and KGW using hard group partitioning settings with different $\gamma$ values on the English C4 dataset.}
    \label{fig:hard_c4}
\end{figure}

Figure~\ref{fig:hard_c4} reports the results under full $\gamma$ sweeps. Since $\gamma$ has different meanings in the two frameworks, we vary it separately according to each definition: in KGW, $\gamma$ denotes the green-token ratio, whereas in TripPattern, it denotes the neutral-group ratio. All other settings follow the main experiments.

\begin{figure}[t!]
\centering
    \includegraphics[width=0.45\textwidth]{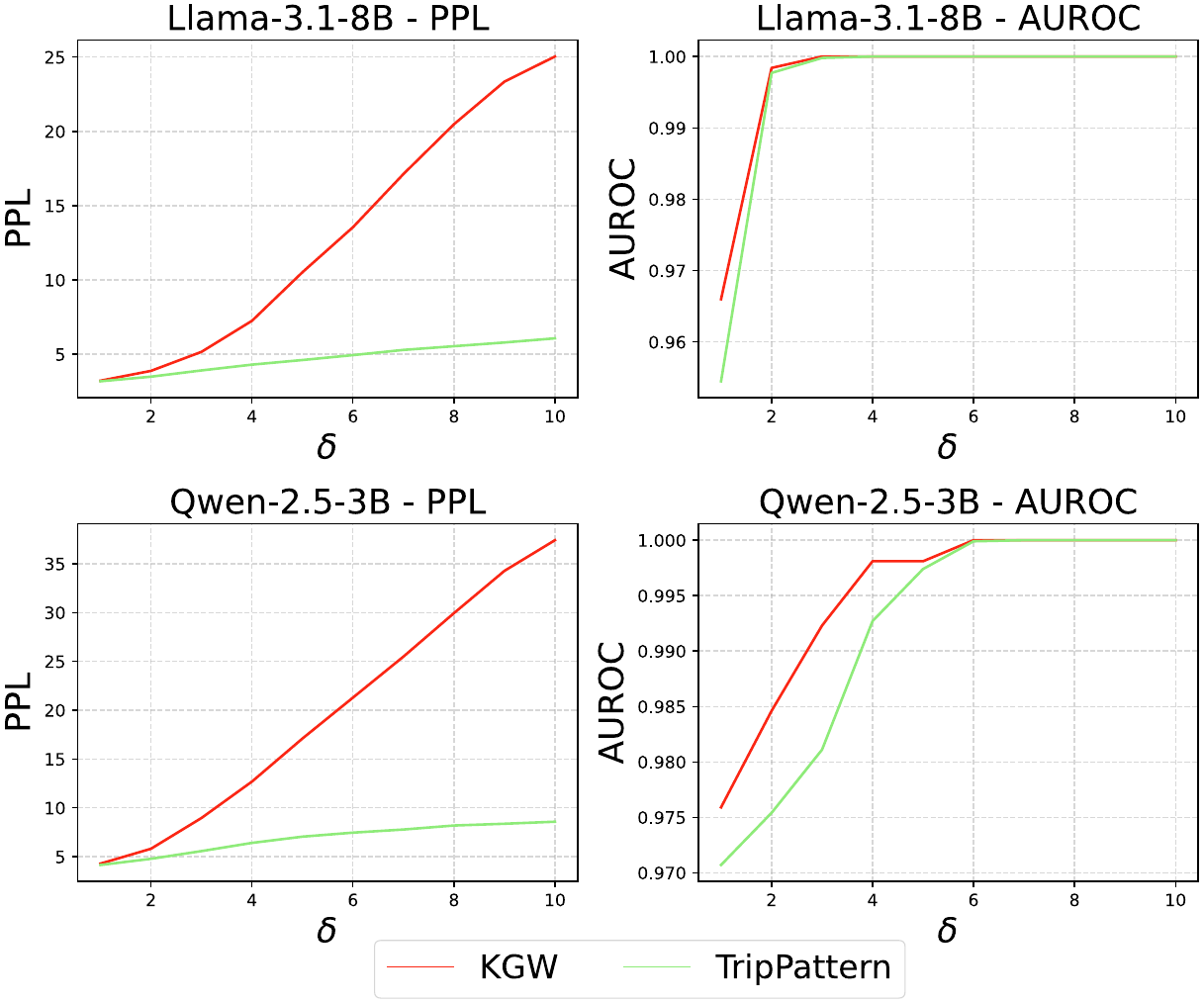}
    \caption{Comparison of TripPattern and KGW with different $\delta$ on the English C4 dataset.}
    \label{fig:kgwVSpattern}
\end{figure}

TripPattern consistently outperforms both TripPattern w/o Len and KGW in text quality, demonstrating its greater token-selection flexibility and the benefit of subword length-based neutral group construction. While all methods achieve similar AUC scores, TripPattern shows a sharp performance drop at large $\gamma$ values, where the neutral group becomes excessively large and weakens the watermark signal.
We further compare TripPattern and KGW by varying the watermark strength $\delta$ on English C4. As shown in Figure~\ref{fig:kgwVSpattern}, TripPattern preserves text quality more effectively than KGW across the $\delta$ sweep while maintaining reliable watermark detection.

\begin{figure*}[t]
\centering
    \includegraphics[width=0.95\textwidth]{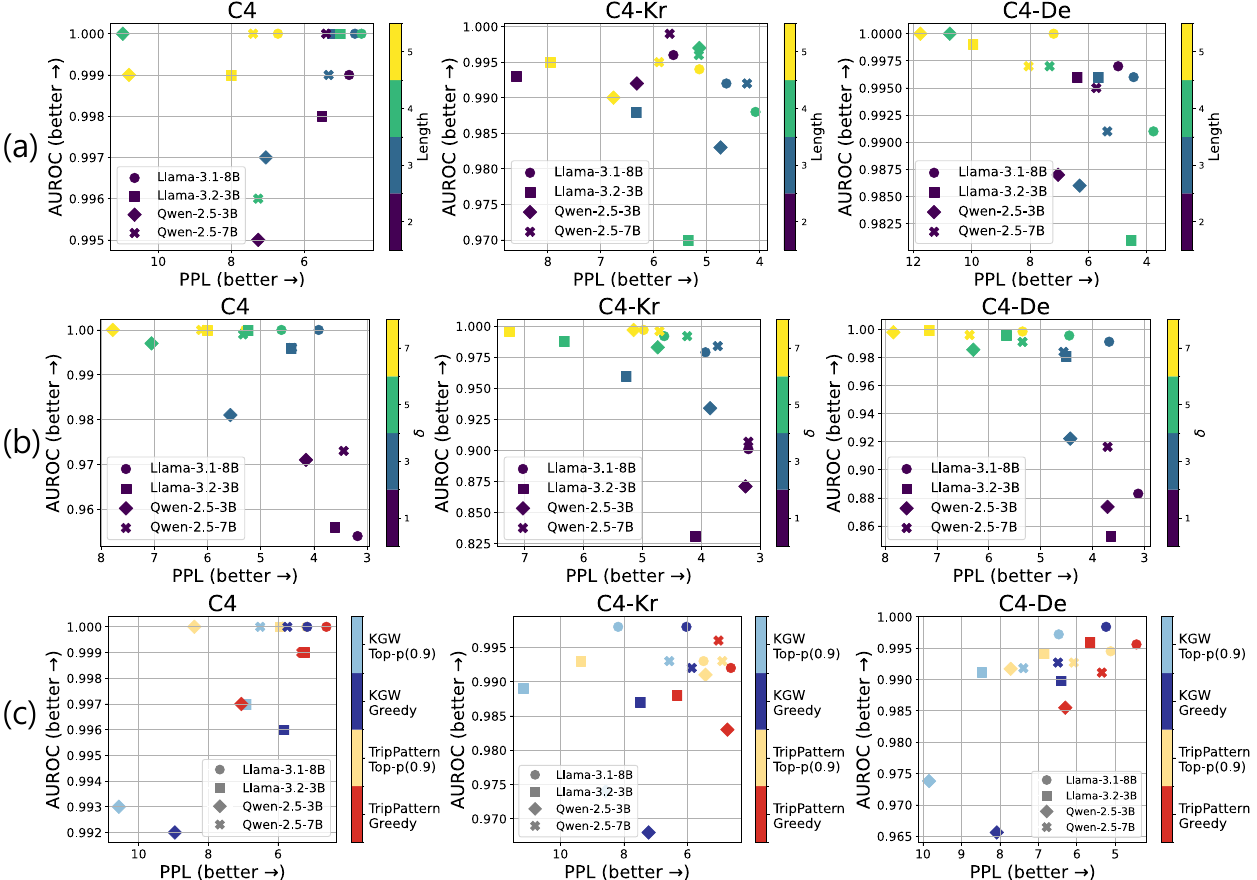}
    \caption{Analysis of key TripPattern design choices. (a) Effect of varying the subword-length threshold for neutral group construction. (b) Effect of varying the watermark strength $\delta$ on text quality (PPL) and detection performance (AUC). (c) Effect of different decoding strategies on text quality and detection performance.}
    \label{fig:merge_analysis}
\end{figure*}

\noindent \textbf{Analysis of Subword Length.}
We examine whether the subword-length threshold used to construct the neutral group is consistent across models and languages. Figure~\ref{fig:merge_analysis} (a) shows that a threshold of three generally provides the best balance between text quality and watermark detectability. In particular, this setting consistently performs well for English, Korean, and German across models. Although increasing the neutral group can improve fluency by allowing more frequent short tokens to remain unconstrained, it may also weaken the watermark signal, revealing a trade-off between text quality and detection performance.

\noindent \textbf{Performance Trade-offs at Different $\delta$.}
We further analyze the effect of the watermark strength $\delta$, which controls the bias applied during token selection~\cite{chen-etal-2024-watme}. As shown in Figure~\ref{fig:merge_analysis} (b), larger $\delta$ values improve watermark detectability but can degrade text quality by imposing stronger constraints on generation. This confirms the expected trade-off between fluency and detection performance.

\noindent \textbf{Effect of Decoding Strategy.}
We also evaluate whether different decoding strategies affect watermark embedding and detection. Figure~\ref{fig:merge_analysis} (c) shows that detection performance remains largely stable across decoding strategies, indicating that TripPattern is robust to decoding choices.

\noindent \textbf{Performance on Text Quality.}
To evaluate text quality beyond PPL, we conducted experiments on CNN/DM~\cite{10.5555/2969239.2969428} and GovReport~\cite{huang-etal-2021-efficient} for long-document summarization. We used ROUGE-L (R-L)~\cite{lin-2004-rouge} and BERTScore (BS)~\cite{zhang2020bertscoreevaluatingtextgeneration} as automatic metrics. We also conducted human evaluation. We sampled 50 instances from CNN/DM~\cite{zhou-etal-2020-level,wang-etal-2023-element}, and using MTurk, 50 evaluators with U.S. high school and bachelor's degrees rated outputs from 1 to 5 (5 is the best) for informativeness (Info.), conciseness (Conc.), and factual consistency (Faith.). Table~\ref{tab:cnndm} and~\ref{tab:humaneval} show the results. * denotes statistically significant (\textit{p}$<$0.01) improvements, compared to the underlined scores. We used paired bootstrap resampling with 100,000 random samples~\cite{koehn-2004-statistical} for the significance test. Our TripPattern generates informative, concise, and faithful summaries while enabling watermark detection.

\begin{table}[t]
\renewcommand{\arraystretch}{1.0}
\centering

\resizebox{0.48\textwidth}{!}{
\begin{tabular}{lcccccccc}\toprule
\multirow{2}{*}{\textbf{Method}} & \multicolumn{4}{c}{\textbf{CNN/DM}} & \multicolumn{4}{c}{\textbf{GovReport}} \\
\cmidrule(lr){2-5} \cmidrule(lr){6-9}
 & \textbf{R-L $\uparrow$} & \textbf{BS $\uparrow$} & \textbf{AUC $\uparrow$} & \textbf{TPR@5 $\uparrow$} &  \textbf{R-L $\uparrow$} & \textbf{BS $\uparrow$} & \textbf{AUC $\uparrow$} & \textbf{TPR@5 $\uparrow$}\\
\midrule
Llama-3.1-8B  & 34.07 & 67.36 & - & - & 27.39 & 56.56 & - & -\\
 +KGW         & \textbf{32.68} & 66.55 & 0.997 & 0.980 & 12.37 & 46.23 & 0.999 & 0.992\\
 +TripPattern & 32.03 & \textbf{66.56} & \textbf{0.999} & \textbf{0.996} & \textbf{30.54} & \textbf{61.08} & \textbf{1.000} & \textbf{1.000}\\
\midrule
Qwen-2.5-3B    & 30.03 & 65.09 & - & - & 25.88 & 59.57 & - & -\\
 +KGW          & 26.77 & 63.50 & \textbf{1.000} & \textbf{1.000} & 20.28 & 56.78 & \textbf{1.000} & \textbf{1.000}\\
 +TripPattern  & \textbf{26.92} & \textbf{63.55} & \textbf{1.000} & \textbf{1.000} & \textbf{23.68} & \textbf{58.94} & \textbf{1.000} & \textbf{1.000}\\
\bottomrule
\end{tabular}}
\caption{Performance on the CNN/DM and GovReport datasets.}
\label{tab:cnndm}
\end{table}

\begin{table}[t]
\renewcommand{\arraystretch}{0.8}
\centering

\resizebox{0.40\textwidth}{!}{
\begin{tabular}{ccccc}\toprule
\textbf{Model} & \textbf{Method} & \textbf{Info.} & \textbf{Conc.} & \textbf{Faith.} \\
\midrule 
\multirow{2}{*}{\shortstack[c]{Qwen-2.5\\-7B-Instruct}} 
& KGW & \underline{3.44} & \underline{3.50} & \underline{3.08} \\
& TripPattern & \textbf{4.00}* & \textbf{3.88}* & \textbf{3.58}* \\
\bottomrule
\end{tabular}}
\caption{Human evaluation results.}
\label{tab:humaneval}
\end{table}

\begin{table}[h]
\renewcommand{\arraystretch}{0.7}
\centering

\resizebox{0.48\textwidth}{!}{
\begin{tabular}{lcccccccc}
\toprule
\multirow{2}{*}{\textbf{Method}} & \multirow{2}{*}{\textbf{Original}} & \multicolumn{3}{c}{\textbf{WordDel}} & \multicolumn{3}{c}{\textbf{WordSub}} & \multirow{2}{*}{\textbf{Dipper}} \\
\cmidrule(lr){3-5} \cmidrule(lr){6-8}
 &  & 0.1 & 0.2 & 0.3 & 0.1 & 0.2 & 0.3 & \\
\midrule
KGW          & 0.992 & 0.992 & 0.993 & \textbf{0.991} & 0.992 & 0.992 & 0.988 & 0.893 \\
Unigram      & 0.990 & 0.987 & 0.986 & 0.985 & 0.986 & 0.984 & 0.981 & \textbf{0.958} \\
UPV          & 0.991 & 0.987 & 0.983 & 0.980 & 0.991 & 0.991 & \textbf{0.991} & 0.873 \\
SWEET        & 0.988 & 0.988 & 0.987 & 0.989 & 0.989 & 0.986 & 0.988 & 0.911 \\
\midrule
TripPattern  & \textbf{0.997} & \textbf{0.997} & \textbf{0.996} & 0.976 & \textbf{0.997} & \textbf{0.995} & 0.951 & 0.675 \\
TripPattern w/o Len & 0.994 & 0.992 & 0.988 & 0.966 & 0.992 & 0.983 & 0.934 & 0.641 \\
\bottomrule
\end{tabular}
}
\caption{AUC under attacks on C4.}
\label{tab:attack-wrap}

\end{table}

\noindent \textbf{Robustness against Attacks.}
We evaluate robustness against three attack types: WordDel, WordSub, and Dipper. WordDel randomly removes a given percentage of words, while WordSub replaces a given percentage of words with WordNet synonyms. Dipper performs context-aware paraphrasing by controlling lexical diversity and content reordering~\cite{krishna2023paraphrasingevadesdetectorsaigenerated}.

Table~\ref{tab:attack-wrap} reports the results using texts generated by Qwen-2.5-3B. TripPattern maintains reasonable detection performance under word-level perturbations, although it slightly underperforms KGW under 30\% WordDel and WordSub attacks. In contrast, TripPattern is more vulnerable to Dipper, likely because context-aware paraphrasing disrupts the local alternation patterns used for detection.

\noindent \textbf{Investigating Softmax Probability.}
To better understand the AUC degradation observed for both KGW and TripPattern on multilingual datasets in Table~\ref{tab:c4_multilingual_full_results}, we measure the average top-1 next-token softmax probability without watermarking. As shown in Table~\ref{tab:softmaxx}, multilingual datasets exhibit substantially higher next-token certainty than English. This suggests that the model distribution is more concentrated in multilingual settings, making generation less responsive to the watermarking bias $\delta$. Consequently, watermark insertion becomes weaker, leading to reduced detection performance.

We provide additional case studies of TripPattern outputs in Appendix~\ref{appen:case}.

\begin{table}[t!]
\renewcommand{\arraystretch}{0.95}
\centering

\resizebox{0.45\textwidth}{!}{
\begin{tabular}{lccccccc}
\toprule
 \textbf{Model} &  \textbf{English} & \textbf{Korean} & \textbf{German} & \textbf{Spanish} \\\midrule
Llama-3.2-3B &  0.67 & 0.75 & 0.73 & 0.72 \\
Qwen-2.5-3B &  0.54 & 0.66 & 0.65 & 0.67 \\
\bottomrule
\end{tabular}}
\caption{Average softmax probabilities on the mC4 datasets.}
\label{tab:softmaxx}
\end{table}

\section{Related Work}
Due to recent advances in LLMs, which can generate fluent human-level text, early post-hoc AI-generated text detection methods formulated as binary classification have become less effective~\cite{jawahar-etal-2020-automatic,10.5555/3618408.3619446}. Another line of work injects watermarks by modifying textual content or structure through rule-based methods~\cite{10.5555/555654} or neural encoders~\cite{yang2021tracingtextprovenancecontextaware,ueoka-etal-2021-frustratingly}. These methods embed messages that can later be extracted using corresponding rules or neural decoders, but substantial modifications often degrade text quality due to the discrete nature of natural language.

Recent watermarking methods mainly operate at the decoding stage by adjusting token logits during generation without modifying the underlying LLM. The KGW framework introduced green/red vocabulary partitioning and green-token biasing~\cite{kirchenbauer2024watermarklargelanguagemodels}, which has since been extended through context-aware partitioning, robust detection rules, mutual exclusivity constraints, fixed green lists, and unbiased output-distribution adjustments~\cite{guo-etal-2024-context-aware,ren-etal-2024-robust,wu-chandrasekaran-2024-bypassing,chen-etal-2024-watme,zhao2024provable,kuditipudi2024robust,hu2024unbiased}.

Despite these advances, existing work has paid limited attention to vocabulary-partitioning watermarking in multilingual settings. To address this gap, we propose a three-group vocabulary partitioning method with two pattern groups and one neutral group. We further incorporate subword token length into neutral group construction, improving text quality and making the watermarking scheme more suitable for multilingual scenarios.

\section{Conclusion}
This study introduced TripPattern, a novel watermarking approach for LLMs that leverages a three-group vocabulary partitioning framework. By explicitly alternating token selection between two distinct pattern groups and incorporating a neutral group, TripPattern significantly enhances the token selection flexibility, preserving text quality and naturalness across multilingual datasets. We further propose considering subword token length as a criterion to define the neutral group. Because little attention has been paid to embedding watermarks in multilingual settings, we evaluate four languages. Our analysis and experimental results show that TripPattern provides a favorable balance between text quality and watermark detectability compared with conventional two-group partitioning methods in multilingual settings.

\section*{Limitations}
Although TripPattern improves the balance between text quality and watermark detectability, it has several limitations. First, TripPattern relies on local alternation patterns between two pattern groups. As shown in our robustness analysis, this makes the method vulnerable to strong context-aware paraphrasing attacks such as Dipper, which can disrupt local token-level patterns while preserving the original meaning. Improving robustness against semantic paraphrasing remains an important direction for future work.

Second, TripPattern still involves a trade-off between text quality and detection performance. The watermark strength $\delta$ and the neutral-group ratio $\gamma$ directly affect this balance. Larger $\delta$ values improve detectability but may reduce generation quality, while larger neutral groups provide greater token-selection flexibility but can dilute the watermark signal. Although we use fixed hyperparameters across languages, optimal settings may vary depending on the model, language, domain, and decoding strategy.

Third, detection performance can degrade in multilingual settings where the model distribution is highly concentrated. Our softmax probability analysis suggests that higher next-token certainty makes generation less responsive to watermarking bias, weakening watermark insertion. This issue may be more pronounced for low-resource languages, domain-specific text, or models with different tokenization behavior.

Finally, our experiments focus on instruction-tuned Llama and Qwen models and evaluate a limited set of languages and tasks. While TripPattern shows strong results across multilingual generation, reasoning, and summarization settings, broader evaluation on more languages, model families, domains, and real-world deployment scenarios is needed to fully assess its generalizability.

\bibliography{custom}

\appendix

\section{TripPattern Algorithm}\label{appen:alg}

\noindent \textbf{Generation Algorithm.}
Algorithm~\ref{alg:TripPattern} presents the generation procedure of TripPattern.
Line~1 initializes the alternation state variable $s_{prev}$, which stores the label of the most recent non-neutral pattern group.
Unlike storing the pattern group set itself, storing only the group label avoids ambiguity because the vocabulary is repartitioned at every generation step.

At each generation step, the vocabulary $\mathcal{V}$ is deterministically repartitioned into a neutral group $\mathcal{V}_n$ and two pattern groups, $\mathcal{V}_{p1}$ and $\mathcal{V}_{p2}$ using the step-specific hash value and PRNG.
During this repartitioning process, short subword tokens are prioritized for assignment to the neutral group until the target neutral-group proportion $\gamma$ is reached, and the remaining tokens are evenly divided between the two pattern groups.
The same deterministic partitioning procedure is used during detection, allowing the detector to reconstruct the corresponding step-specific partitions.

Lines~3--4 compute the step-specific hash value and reconstruct the current vocabulary partition.
Lines~5--11 determine the current target pattern group according to the alternation state $s_{prev}$.
If the most recent non-neutral pattern token was assigned to group 1, the current target group is set to $\mathcal{V}_{p2}$.
If it was assigned to group 2, the current target group is set to $\mathcal{V}_{p1}$.
If no previous non-neutral pattern token has been observed, $\mathcal{V}_{p1}$ is used as the default initial target group.

Line~12 defines the allowable token set $\mathcal{A}_t$ as the union of the selected target pattern group and the neutral group.
Lines~13--16 obtain the next-token logits from the model, add the watermark strength $\delta$ to the logits of tokens in $\mathcal{A}_t$, normalize the adjusted logits over the full vocabulary, and sample the next token.
This soft-bias procedure statistically encourages the model to select tokens from the current target pattern group or the neutral group while preserving the possibility of sampling from the full vocabulary.

Lines~17--23 update the alternation state after sampling.
If the sampled token belongs to $\mathcal{V}_{p1}$, the state is updated to $s_{prev}=1$.
If the sampled token belongs to $\mathcal{V}_{p2}$, the state is updated to $s_{prev}=2$.
If the sampled token belongs to the neutral group, the state remains unchanged, since neutral tokens do not carry watermark signals.
By repeating this process, TripPattern embeds a detectable alternation pattern between the two pattern groups while allowing neutral tokens to preserve generation flexibility and fluency.

\begin{algorithm}[t!]
\caption{Text Generation with TripPattern}
\small
\label{alg:TripPattern}
\begin{algorithmic}[1]
\REQUIRE Prompt tokens $\mathbf{x}_{1:n} = \{x_1, x_2, \dots, x_n\}$, neutral group proportion $\gamma \in (0,1)$, watermark strength $\delta > 0$, model $\mathcal{M}$, vocabulary $\mathcal{V}$
\ENSURE Watermarked text $x_{n+1}, x_{n+2}, \dots, x_{n+T}$

\STATE Initialize the alternation state $s_{prev}\leftarrow\varnothing$, where $s_{prev}\in\{1,2,\varnothing\}$ stores the label of the most recent non-neutral pattern group.

\FOR{$t = 1, 2, \dots, T$}
    \STATE $h_{n+t}\leftarrow H(x_{n+t-1})$
    
    \STATE Reconstruct the step-specific partition
    $(\mathcal{V}_{n}, \mathcal{V}_{p1}, \mathcal{V}_{p2})$
    using $h_{n+t}$, the PRNG, neutral-group proportion $\gamma$, and the subword-length priority rule.

    \IF{$s_{prev}=1$}
        \STATE Set current target pattern group $\mathcal{V}_{p} \leftarrow \mathcal{V}_{p2}$.
    \ELSIF{$s_{prev}=2$}
        \STATE Set current target pattern group $\mathcal{V}_{p} \leftarrow \mathcal{V}_{p1}$.
    \ELSE
        \STATE Set current target pattern group $\mathcal{V}_{p} \leftarrow \mathcal{V}_{p1}$.
    \ENDIF

    \STATE Allowable token set: $\mathcal{A}_t = \mathcal{V}_{p} \cup \mathcal{V}_n$.

    \STATE Obtain logits $l^{n+t} \in \mathbb{R}^{|\mathcal{V}|}$ from model $\mathcal{M}$.

    \STATE Adjust logits by adding watermark strength $\delta$ to tokens in the allowable token set $\mathcal{A}_{t}$:
    \[
    l_i^{n+t} =
    \begin{cases}
    l_i^{n+t} + \delta, & \text{if } v_i \in \mathcal{A}_{t},\\[6pt]
    l_i^{n+t}, & \text{otherwise}.
    \end{cases}
    \]

    \STATE Compute probability distribution for token selection:
    \[
    \hat{p}_{n+t}[i] =
    \frac{\exp(l_i^{n+t})}{\sum_{j \in \mathcal{V}} \exp(l_j^{n+t})},
    \quad \forall v_i \in \mathcal{V}.
    \]

    \STATE Sample the next token $x_{n+t}$ from distribution $\hat{p}_{n+t}$.

    \IF{$x_{n+t} \in \mathcal{V}_{p1}$}
        \STATE Update alternation state $s_{prev}\leftarrow 1$.
    \ELSIF{$x_{n+t} \in \mathcal{V}_{p2}$}
        \STATE Update alternation state $s_{prev}\leftarrow 2$.
    \ELSE
        \STATE Keep $s_{prev}$ unchanged.
    \ENDIF

\ENDFOR

\STATE \textbf{return} Generated tokens: $x_{n+1}, x_{n+2}, \dots, x_{n+T}$.
\end{algorithmic}
\end{algorithm}

\noindent \textbf{Detection Algorithm.}
Algorithm~\ref{alg:TripPatterndetect} describes the detection procedure of TripPattern.
The detector receives a token sequence $\mathbf{x}_{0:n}$, where $x_0$ is either the preceding prompt token or a predefined initial seed token.
This makes the hash input $H(x_{i-1})$ well-defined for the first detected token.
Line~1 initializes an empty label list $\mathbf{y}$.

For each token position, the detector reconstructs the position-specific partition of the vocabulary into a neutral group $\mathcal{V}_n$ and two pattern groups, $\mathcal{V}_{p1}$ and $\mathcal{V}_{p2}$ using the same hash/PRNG procedure, neutral-group proportion $\gamma$, and subword-length priority rule as in generation.
Lines~4--10 map each non-neutral token to its pattern label: tokens in $\mathcal{V}_{p1}$ are assigned \quotes{1}, and tokens in $\mathcal{V}_{p2}$ are assigned \quotes{2}.
Neutral tokens are excluded because they do not carry watermark signals.

After obtaining the pattern-label sequence, the detector computes the number of adjacent pattern-token pairs and the number of alternations $C$ between adjacent labels.
It then calculates the observed alternation ratio $r=C/m$ and compares it against the permutation null conditional on $(n_1,n_2)$, where $n_1$ and $n_2$ are the numbers of labels \quotes{1} and \quotes{2}, respectively.
The z-score is computed using the null moments of the runs statistic.
If the z-score exceeds a predefined threshold, the null hypothesis that the text does not follow the TripPattern alternation rule is rejected, and the watermark is identified.

\begin{algorithm}[t!]
\caption{Text Detection with TripPattern}
\small
\label{alg:TripPatterndetect}
\begin{algorithmic}[1]
\REQUIRE Text tokens $\mathbf{x}_{0:n} = \{x_0, x_1, \dots, x_n\}$, where $x_0$ is a prompt token or a predefined initial seed token; neutral group proportion $\gamma \in (0,1)$; vocabulary $\mathcal{V}$
\ENSURE z-score ($z$)

\STATE Initialize an empty list $\mathbf{y} \leftarrow []$

\FOR{$i = 1, 2, \dots, n$}
    \STATE $h_i \leftarrow H(x_{i-1})$

    \STATE Reconstruct the position-specific partition
    $(\mathcal{V}_{n}, \mathcal{V}_{p1}, \mathcal{V}_{p2})$
    using $h_i$, the PRNG, neutral-group proportion $\gamma$, and the subword-length priority rule.

    \IF{$x_i \in \mathcal{V}_{p1}$}
        \STATE Append \quotes{1} to $\mathbf{y}$
    \ELSIF{$x_i \in \mathcal{V}_{p2}$}
        \STATE Append \quotes{2} to $\mathbf{y}$
    \ENDIF
\ENDFOR

\STATE Let $k = |\mathbf{y}|$ be the number of non-neutral pattern tokens.

\IF{$k < 2$}
    \STATE \textbf{return} $z=0$
\ENDIF

\STATE Let \(m = k - 1\) be the number of adjacent pattern-token pairs.

\STATE Compute the number of alternations between adjacent pattern labels:
\[
C = \sum_{i=1}^{m} \mathbb{I}[y_i \neq y_{i+1}].
\]

\STATE Calculate the observed alternation ratio:
\[
r = \frac{C}{m}.
\]

\STATE Let $n_1=\sum_{i=1}^{k}\mathbb{I}[y_i=1]$ and $n_2=k-n_1$.

\STATE Under the permutation null conditional on $(n_1,n_2)$,
\[
\mu_R = 1+\frac{2n_1n_2}{k},
\qquad
\sigma_R^2 =
\frac{2n_1n_2(2n_1n_2-k)}{k^2(k-1)}.
\]

\IF{$\sigma_R^2 = 0$}
    \STATE \textbf{return} $z=0$
\ENDIF

\STATE Since $R=C+1$ and $r=C/(k-1)$,
\[
\mathbb{E}[r]
=
\frac{\mu_R-1}{k-1}
=
\frac{2n_1n_2}{k(k-1)},
\qquad
\mathrm{Var}(r)
=
\frac{\sigma_R^2}{(k-1)^2}.
\]

\STATE Compute the z-score:
\[
z =
\frac{r-\mathbb{E}[r]}{\sqrt{\mathrm{Var}(r)}}
=
\frac{R-\mu_R}{\sigma_R}.
\]

\STATE \textbf{return} z-score $z$.
\end{algorithmic}
\end{algorithm}

\section{Proof}\label{appen:proof}
We provide a mathematical analysis demonstrating how TripPattern outperforms the KGW-based framework, the commonly used two-group partitioning framework of \quotes{green} and \quotes{red} tokens, focusing on the flexibility of token selection at each decoding step.

\noindent \textbf{Definition (Vocabulary Partition).} \textit{Let $\textbf{V}$ be the vocabulary of the language model, and let $\textbf{|V|}$ represent its size. Under the KGW-based framework, the vocabulary is partitioned into two groups, $\textbf{V}=\textbf{V$_g$} \cup \textbf{V$_r$}, \textbf{V$_g$} \cap \textbf{V$_r$} = \emptyset$. Here, $\textbf{V$_g$}$ and $\textbf{V$_r$}$ denote the set of \quotes{green} and \quotes{red} tokens. Under the TripPattern framework, the vocabulary is partitioned into three groups, $\textbf{V}=\textbf{V$_{p1}$} \cup \textbf{V$_{p2}$} \cup \textbf{V$_n$}, \textbf{V$_{p1}$} \cap \textbf{V$_{p2}$} = \textbf{V$_{p1}$} \cap \textbf{V$_n$} = \textbf{V$_{p2}$} \cap \textbf{V$_n$} = \emptyset$. Here, \textbf{V$_{p1}$} and \textbf{V$_{p2}$} denote the two pattern groups and \textbf{V$_n$} is the neutral group.}

\noindent \textbf{Proof.} Consider an arbitrary decoding step \textit{\textbf{t}} during text generation. Let \textit{\textbf{$A_t$}} denote the set of allowable tokens for selection at each decoding step \textit{\textbf{t}}. At each decoding step, KGW methods consistently bias selection towards tokens in the green group, \textit{\textbf{$V_g$}}, resulting in a fixed allowable selection set: 
\begin{equation}
    |\mathbf{A}_t^{KGW}| = |\mathbf{V}_g|,
\end{equation}
In contrast, at each decoding step in TripPattern, the allowable selection set \textit{$\mathbf{A}_t$}, dynamically depends on the pattern group of the previously generated token, $\mathbf{V}_{prev}$, as defined below:

\begin{equation}
\mathbf{V}_p \!\!= \!\!\left\{
\begin{aligned}
&\!\mathbf{V}_{p2}, &&\!\!\!\!\!\text{if $\mathbf{V_{prev}}$=}\ \mathbf{V}_{p1}\\
&\!\mathbf{V}_{p1}, &&\!\!\!\!\!\text{if $\mathbf{V_{prev}}$=}\ \mathbf{V}_{p2}\\
&\!(\mathbf{V}_{p1} \!\cup\! \mathbf{V}_{p2}) \!-\! \mathbf{V}_{prev}, &&\!\!\!\!\!\text{otherwise}
\end{aligned}
\right.
\end{equation}
where the initial $\mathbf{V_{prev}}$ is set to \textit{None}, and the allowable selection set for TripPattern is given by:
\begin{equation}
    \mathbf{A}_{t}^{TripPattern}=\mathbf{V}_{p} \ \cup \ \mathbf{V}_n.
\end{equation}

Given that the neutral group \textit{\textbf{$V_n$}} always increases the allowable selection set compared to the two-group method, thus it directly follows:
\begin{equation}
    |\mathbf{A}_t^{TripPattern}| \ge |\mathbf{A}^{KGW}_t|
\end{equation}
when the neutral group \textit{$\mathbf{V}_n$} is non-empty and sufficiently large. 

Thus, TripPattern mathematically ensures greater flexibility in token selection than KGW-based methods (when $\gamma_{\mathrm{KGW}} \le \frac{1+\gamma_{N}}{2}$), directly enhancing the naturalness and quality of generated texts. Here, $\gamma_{KGW}$ indicates the proportion of green tokens for KGW-based methods and $\gamma_{N}$ indicates the proportion of neutral tokens for TripPattern.

\begin{table*}[t!]
\caption{Experimental results for Morphmark and TripPattern on the C4 and mC4 datasets}
\label{tab:c4_morphmark}
\renewcommand{\arraystretch}{0.9}
\centering
\resizebox{0.95\textwidth}{!}{
\begin{tabular}{lccccccccccccccc}
\toprule
 & \multicolumn{3}{c}{\textbf{English}} & \multicolumn{3}{c}{\textbf{Korean}}  & \multicolumn{3}{c}{\textbf{German}} & \multicolumn{3}{c}{\textbf{Spanish}} & \multicolumn{3}{c}{\textbf{Average}} \\
\cmidrule(lr){2-4} \cmidrule(lr){5-7} \cmidrule(lr){8-10} \cmidrule(lr){11-13} \cmidrule(lr){14-16}
\multirow{-2}{*}{\textbf{Model}} & \textbf{PPL $\downarrow$} & \textbf{AUC $\uparrow$} & \textbf{TPR@5 $\uparrow$} & \textbf{PPL $\downarrow$} & \textbf{AUC $\uparrow$} & \textbf{TPR@5 $\uparrow$} & \textbf{PPL $\downarrow$} & \textbf{AUC $\uparrow$} & \textbf{TPR@5 $\uparrow$} & \textbf{PPL $\downarrow$} & \textbf{AUC $\uparrow$} & \textbf{TPR@5 $\uparrow$} & \textbf{PPL $\downarrow$} & \textbf{AUC $\uparrow$} & \textbf{TPR@5 $\uparrow$} \\
\midrule
Llama-3.2-3B  & 3.49 & - & - & 3.86 & - & - & 3.53 & - & - & 3.14 & - & - & 3.51 & - & - \\
\cline{2-16}
\addlinespace[0.2em] 
 +Morphmark & \textbf{3.62} & 0.974 & 0.852 & \textbf{4.65} & 0.898 & 0.672 & \textbf{3.86} & 0.924 & 0.660 & \textbf{3.35} & 0.934 & 0.744 & \textbf{3.87} & 0.933 & 0.732 \\
 \addlinespace[0.1em]            
 \cdashline{2-16}
 \addlinespace[0.3em]
 +TripPattern         & 5.24 & \textbf{1.000} & \textbf{1.000} & 6.33 & \textbf{0.995} & 0.968 & 5.65 & \textbf{0.996} & \textbf{0.982} & 4.53 & 0.988 & 0.968 & 5.44 & \textbf{0.995} & 0.980 \\
 +TripPattern w/o Len & 5.26 & 0.998 & 0.990 & 7.88 & 0.986 & \textbf{0.974} & 6.00 & 0.990 & 0.972 & 5.18 & \textbf{0.996} & \textbf{0.988} & 6.08 & 0.993 & \textbf{0.981}\\
\midrule
Llama-3.1-8B  & 3.08 & - & - & 3.06 & - & - & 3.07 & - & - & 2.66 & - & - & 2.97 & - & - \\
\cline{2-16}
\addlinespace[0.2em] 
 +Morphmark & \textbf{3.23} & 0.971 & 0.892 & \textbf{3.35} & 0.928 & 0.742 & \textbf{3.25} & 0.937 & 0.676 & \textbf{2.81} & 0.937 & 0.730 & \textbf{3.16} & 0.943 & 0.760 \\
\addlinespace[0.1em]            
 \cdashline{2-16}
 \addlinespace[0.3em]
 +TripPattern         & 4.61 & \textbf{1.000} & \textbf{1.000} & 4.63 & 0.990 & 0.970 & 4.45 & 0.993 & 0.986 & 3.63 & 0.995 & 0.992 & 4.33 & 0.995 & 0.987 \\
 +TripPattern w/o Len & 4.61 & 0.999 & 0.996 & 5.23 & \textbf{0.999} & \textbf{0.996} & 4.80 & \textbf{1.000} & \textbf{0.999} & 3.88 & \textbf{1.000} & 0.999 & 4.63 & \textbf{1.000} & \textbf{0.998} \\
\midrule
Qwen-2.5-3B  & 3.93 & - & - & 3.30 & - & - & 3.68 & - & - & 3.19 & - & - & 3.53 & - & - \\
\cline{2-16}
\addlinespace[0.2em] 
 +Morphmark & \textbf{4.29} & 0.968 & 0.952 & \textbf{3.41} & 0.887 & 0.724 & \textbf{3.75} & 0.893 & 0.748 & \textbf{2.96} & 0.847 & 0.672 & \textbf{3.60} & 0.899 & 0.774 \\
 \addlinespace[0.1em]            
 \cdashline{2-16}
 \addlinespace[0.3em]
 +TripPattern         & 7.05 & \textbf{0.997} & \textbf{0.994} & 4.74 & \textbf{0.985} & 0.956 & 6.30 & 0.986 & 0.964 & 4.91 & \textbf{0.969} & 0.924 & 5.75 & \textbf{0.984} & 0.960\\
 +TripPattern w/o Len & 6.80 & 0.994 & 0.984 & 6.17 & 0.984 & \textbf{0.962} & 7.04 & \textbf{0.989} & \textbf{0.966} & 5.37 & \textbf{0.969} & \textbf{0.940} & 6.35 & \textbf{0.984} & \textbf{0.963}\\
\midrule
Qwen-2.5-7B  & 3.24 & - & - & 3.17 & - & - & 3.49 & - & - & 2.90 & - & - & 3.20 & - & - \\
\cline{2-16}
\addlinespace[0.2em] 
 +Morphmark & \textbf{3.40} & 0.983 & 0.914 & \textbf{3.39} & 0.957 & 0.838 & \textbf{3.75} & 0.954 & 0.810 & \textbf{3.01} & 0.954 & 0.856 & \textbf{3.39} & 0.962 & 0.855 \\
\addlinespace[0.1em]            
 \cdashline{2-16}
 \addlinespace[0.3em]
 +TripPattern         & 5.33 & 0.999 & 0.996 & 5.02 & 0.992 & \textbf{0.986} & 5.35 & 0.992 & 0.986 & 4.26 & 0.995 & 0.988 & 4.80 & 0.995 & 0.989 \\
 +TripPattern w/o Len & 5.07 & \textbf{1.000} & \textbf{1.000} & 4.24 & \textbf{0.997} & \textbf{0.986} & 5.54 & \textbf{0.997} & \textbf{0.992} & 4.26 & \textbf{0.996} & \textbf{0.996} & 4.97 & \textbf{0.998} & \textbf{0.994}\\
\bottomrule
\end{tabular}}
\end{table*}

\begin{table*}[h]
\caption{Experimental results for Morphmark and TripPattern on the GSM8K and MGSM datasets.}
\label{tab:gsm_morphmark}
\renewcommand{\arraystretch}{0.9}
\centering
\resizebox{0.95\textwidth}{!}{
\begin{tabular}{lccccccccccccccc}
\toprule
 & \multicolumn{3}{c}{\textbf{English}} & \multicolumn{3}{c}{\textbf{Korean}} & \multicolumn{3}{c}{\textbf{German}} & \multicolumn{3}{c}{\textbf{Spanish}} & \multicolumn{3}{c}{\textbf{Average}} \\
\cmidrule(lr){2-4} \cmidrule(lr){5-7} \cmidrule(lr){8-10} \cmidrule(lr){11-13} \cmidrule(lr){14-16}
 \multirow{-2}{*}{\textbf{Model}} & \textbf{ACC $\uparrow$} & \textbf{AUC $\uparrow$} & \textbf{TPR@5 $\uparrow$} & \textbf{ACC $\uparrow$} & \textbf{AUC $\uparrow$} & \textbf{TPR@5 $\uparrow$} & \textbf{ACC $\uparrow$} & \textbf{AUC $\uparrow$} & \textbf{TPR@5 $\uparrow$} & \textbf{ACC $\uparrow$} & \textbf{AUC $\uparrow$} & \textbf{TPR@5 $\uparrow$} & \textbf{ACC $\uparrow$} & \textbf{AUC $\uparrow$} & \textbf{TPR@5 $\uparrow$}\\
\midrule
Llama-3.2-3B  & 72.4 & - & - & 7.35 & - & - & 11.6 & - & - & 11.45 & - & - & 25.70 & - & - \\
\cline{2-16}
\addlinespace[0.2em] 
 +Morphmark & \textbf{72.25} & 0.750 & 0.253 & 7.43 & 0.745 & 0.272 & \textbf{11.37} & 0.715 & 0.156 & \textbf{11.45} & 0.718 & 0.252 & \textbf{25.63} & 0.732 & 0.233 \\
 \addlinespace[0.1em]            
 \cdashline{2-16}
 \addlinespace[0.3em]
 +TripPattern         & 67.10 & 0.950 & 0.776 & \textbf{7.96} & 0.965 & 0.848 & 10.54 & \textbf{0.966} & \textbf{0.816} & \textbf{11.45} & 0.932 & 0.648 & 24.26 & 0.953 & 0.772\\
 +TripPattern w/o Len & 64.22 & \textbf{0.964} & \textbf{0.817} & 5.46 & \textbf{0.977} & \textbf{0.892} & 9.78  & 0.946 & 0.784 & 9.78  & \textbf{0.938} & \textbf{0.760} & 22.31 & \textbf{0.956} & \textbf{0.813} \\
\midrule
Llama-3.1-8B  & 81.05 & - & - & 10.31 & - & - & 13.65 & - & - & 13.95 & - & - & 29.74 & - & - \\
\cline{2-16}
\addlinespace[0.2em] 
 +Morphmark & \textbf{78.92} & 0.752 & 0.287 & \textbf{11.22} & 0.751 & 0.356 & \textbf{13.34} & 0.712 & 0.148 & \textbf{13.50} & 0.728 & 0.216 & \textbf{29.25} & 0.736 & 0.252\\
 \addlinespace[0.1em]            
 \cdashline{2-16}
 \addlinespace[0.3em]
 +TripPattern         & 75.74 & 0.944 & 0.740 & 10.46 & 0.951 & 0.776 & 12.59 & \textbf{0.973} & \textbf{0.892} & 12.51 & 0.945 & \textbf{0.760} & 27.83 & 0.953 & 0.792\\
 +TripPattern w/o Len & 69.90 & \textbf{0.966} & \textbf{0.835} & 8.11  & \textbf{0.970} & \textbf{0.840} & 10.92 & 0.959 & 0.848 & 11.83 & \textbf{0.953} & 0.744 & 25.19 & \textbf{0.962} & \textbf{0.817}\\
\midrule
Qwen-2.5-3B  & 33.06 & - & - & 9.25 & - & - & 6.52 & - & - & 7.28 & - & - & 14.03 & - & -\\
\cline{2-16}
\addlinespace[0.2em] 
 +Morphmark & \textbf{35.94} & 0.722 & 0.227 & \textbf{7.88} & 0.840 & 0.504 & \textbf{7.13} & 0.774 & 0.280 & 7.20 & 0.748 & 0.248 & \textbf{14.54} & 0.771 & 0.315\\
 \addlinespace[0.1em]            
 \cdashline{2-16}
 \addlinespace[0.3em]
 +TripPattern         & 31.54 & 0.948 & \textbf{0.768} & 7.81 & 0.980 & 0.904 & 5.46 & \textbf{0.992} & \textbf{0.956} & \textbf{7.43} & \textbf{0.982} & \textbf{0.932} & 13.06 & \textbf{0.976} & \textbf{0.890}\\
 +TripPattern w/o Len & 30.93 & \textbf{0.950} & 0.755 & 6.67 & \textbf{0.988} & \textbf{0.928} & 6.29 & 0.971 & 0.880 & 6.82 & 0.977 & 0.868 & 12.68 & 0.972 & 0.858\\
\midrule
Qwen-2.5-7B  & 55.27 & - & - & 7.51 & - & - & 6.29 & - & - & 6.37 & - & - & 18.86 & - & - \\
\cline{2-16}
\addlinespace[0.2em] 
 +Morphmark & \textbf{54.36} & 0.637 & 0.129 & \textbf{6.82} & 0.718 & 0.260 & \textbf{7.05} & 0.706 & 0.296 & \textbf{7.28} & 0.651 & 0.180 & \textbf{18.88} & 0.678 & 0.216\\
 \addlinespace[0.1em]            
 \cdashline{2-16}
 \addlinespace[0.3em]
 +TripPattern         & 52.24 & \textbf{0.870} & \textbf{0.503} & 6.44 & 0.924 & \textbf{0.716} & 5.99 & \textbf{0.911} & \textbf{0.624} & 5.99 & 0.864 & \textbf{0.536} & 17.67 & \textbf{0.892} & \textbf{0.595}\\
 +TripPattern w/o Len & 41.39 & 0.849 & 0.461 & 6.44 & \textbf{0.935} & 0.712 & 6.52 & 0.890 & 0.560 & 5.99 & \textbf{0.869} & 0.464 & 15.09 & 0.886 & 0.549\\
\bottomrule
\end{tabular}}
\end{table*}

\section{Additional Experimental Results}\label{appen:morphmark}
We additionally compared our TripPattern to MorphMark~\citep{wang-etal-2025-morphmark}, which adaptively scales the bias according to the token-level green probability mass on KGW. Table~\ref{tab:c4_morphmark} and Table~\ref{tab:gsm_morphmark} show the results including MorphMark. TripPattern consistently achieves substantially higher detection performance than MorphMark across models and languages, while maintaining practical generation quality.

\section{Case Study}\label{appen:case}
Table~\ref{tab:casestudy} shows an example of outputs generated by Llama-3.2-3B-Instruct based on TripPattern.

\section{Bin-Wise Analysis}\label{appen:bin}
Table~\ref{tab:bins_auc_tpr} shows the results for the bin-wise analysis on the C4 dataset. We confirm that TripPattern consistently detects even when fewer tokens are generated.

\begin{table*}[h]
\caption{Example of generated texts with TripPattern on the English C4 and GSM datasets. \colorbox{cyan!20}{Blue} and \colorbox{yellow!40}{yellow} represent the two different pattern groups, respectively.}
\label{tab:casestudy}
\centering
\resizebox{0.75\textwidth}{!}{
\begin{tabular}{|c|p{13cm}|c|}
\hline
\textbf{Data} & \parbox{13cm}{\centering \textbf{Watermarked-Text}} & \textbf{Z-score} \\
\hline
C4-English & \parbox{13cm}{and\colorbox{cyan!20}{ that} the \colorbox{yellow!40}{ industry} is \colorbox{cyan!20}{ facing}  a \colorbox{yellow!40}{ perfect storm} of \colorbox{cyan!20}{ declining} \colorbox{yellow!40}{ advertising} \colorbox{yellow!40}{ revenue}, increasing \colorbox{cyan!20}{ production} costs, and the \colorbox{yellow!40}{ rise} of \colorbox{cyan!20}{ digital media}. They \colorbox{yellow!40}{ emphasized} the \colorbox{cyan!20}{ need} for a \colorbox{yellow!40}{ comprehensive} \colorbox{cyan!20}{ approach} to \colorbox{yellow!40}{ address} \colorbox{cyan!20}{ these} \colorbox{yellow!40}{ challenges}, \colorbox{cyan!20}{ rather} \colorbox{yellow!40}{ than} \colorbox{cyan!20}{ simply} \colorbox{yellow!40}{ looking} for \colorbox{cyan!20}{ short-term} cuts. The \colorbox{yellow!40}{ meeting} \colorbox{cyan!20}{ also} \colorbox{yellow!40}{ touched} on the \colorbox{cyan!20}{ importance} of \colorbox{yellow!40}{ divers}\colorbox{cyan!20}{ification} and exploring new \colorbox{yellow!40}{ business} models to \colorbox{cyan!20}{ sustain} the \colorbox{yellow!40}{ company}'s \colorbox{cyan!20}{ long-term} viability. \colorbox{yellow!40}{ While} no \colorbox{yellow!40}{ specific} \colorbox{cyan!20}{ solutions} \colorbox{yellow!40}{ were} \colorbox{cyan!20}{ proposed}, the \colorbox{yellow!40}{ discussions} \colorbox{cyan!20}{ were} \colorbox{yellow!40}{ seen} as a crucial \colorbox{cyan!20}{ step} in \colorbox{yellow!40}{ laying} the \colorbox{cyan!20}{ groundwork} for \colorbox{yellow!40}{ more} \colorbox{cyan!20}{ substantive} \colorbox{yellow!40}{ talks} in the future.

The\colorbox{cyan!20}{ Star}\colorbox{yellow!40}{ Tribune}'s\colorbox{yellow!40}{ financial}\colorbox{cyan!20}{ struggles} are\colorbox{yellow!40}{ well}-d\colorbox{cyan!20}{ocumented},\colorbox{yellow!40}{ with} the paper's\colorbox{cyan!20}{ cash}\colorbox{yellow!40}{ flows}\colorbox{cyan!20}{ consistently} lagging\colorbox{cyan!20}{ behind} its\colorbox{cyan!20}{ debt}\colorbox{yellow!40}{ payments}. The\colorbox{cyan!20}{ company}'s\colorbox{yellow!40}{ debt}-to-equity\colorbox{cyan!20}{ ratio} has become a\colorbox{yellow!40}{ significant concern}, and the\colorbox{cyan!20}{ possibility} of a\colorbox{yellow!40}{ debt}\colorbox{cyan!20}{ restructuring} or\colorbox{yellow!40}{ even} bankruptcy looms\colorbox{yellow!40}{ large}. The Strib's\colorbox{cyan!20}{ decline} has\colorbox{yellow!40}{ also} had a\colorbox{cyan!20}{ knock}-on\colorbox{yellow!40}{ effect} on the\colorbox{cyan!20}{ broader} newspaper\colorbox{yellow!40}{ industry}, as\colorbox{cyan!20}{ investors} and\colorbox{yellow!40}{ lenders}\colorbox{cyan!20}{ become increasingly} cautious\colorbox{yellow!40}{ about} the\colorbox{cyan!20}{ financial} prospects of other media\colorbox{yellow!40}{ outlets}.

The\colorbox{cyan!20}{ fact}\colorbox{yellow!40}{ that} the Star\colorbox{yellow!40}{ Tribune} has\colorbox{cyan!20}{ turned} to\colorbox{yellow!40}{ Restr}uct}
& 6.81\\
\hline
GSM8K-English & 
\parbox{13cm}{
Let's think\colorbox{cyan!20}{ backwards}:\\
The\colorbox{yellow!40}{ final} batch of\colorbox{cyan!20}{ sauce} is 32 oz.\\
The problem states\colorbox{yellow!40}{ that} the\colorbox{cyan!20}{ tomatoes}\colorbox{yellow!40}{ lose} 50\% of\colorbox{cyan!20}{ their} volume \colorbox{yellow!40}{ when}\colorbox{cyan!20}{ cooked}, so we need to find out what the\colorbox{yellow!40}{ original} amount of\colorbox{cyan!20}{ sauce} was before it was cooked.\\
\\
\colorbox{yellow!40}{Since} the volume is halved, we can multiply the amount of\colorbox{cyan!20}{ sauce} by 2 to get the\colorbox{yellow!40}{ original} amount:\\
32 * 2 = 64 oz.\\
\\
Now we\colorbox{cyan!20}{ know} the\colorbox{yellow!40}{ original} amount of\colorbox{cyan!20}{ sauce}.\colorbox{yellow!40}{ Each} 16 oz can of\colorbox{cyan!20}{ sauce} contains 3\colorbox{yellow!40}{ tomatoes}. To find the\colorbox{cyan!20}{ number} of\colorbox{yellow!40}{ tomatoes}, we can divide the amount of\colorbox{cyan!20}{ sauce} by the amount of\colorbox{yellow!40}{ tomatoes} per can:\\
64 / 16 = 4\\
So there were 4\colorbox{cyan!20}{ cans} of\colorbox{yellow!40}{ tomatoes}.\\
Each can has 3\colorbox{yellow!40}{ tomatoes}, so the\colorbox{cyan!20}{ number} of\colorbox{yellow!40}{ tomatoes} is:\\
4 * 3 = 12\\
The\colorbox{yellow!40}{ answer} is 12. \\
\\
However, the\colorbox{cyan!20}{ answer} was not asked, but we were asked how may\colorbox{yellow!40}{ tomatoes}\colorbox{cyan!20}{ were} in the\colorbox{yellow!40}{ cans}\colorbox{cyan!20}{ used}
} & 5.43 \\
\hline
\end{tabular}
}
\end{table*}

\begin{table*}[h]
\caption{Comparison of AUC and TPR@5 across token-length bins.}
\label{tab:bins_auc_tpr}
\centering
\resizebox{0.85\textwidth}{!}{
\begin{tabular}{llcccccccccccccc}
\toprule
\multirow{2}{*}{Model} & \multirow{2}{*}{Method} &
\multicolumn{2}{c}{25--50} &
\multicolumn{2}{c}{50--75} &
\multicolumn{2}{c}{75--100} &
\multicolumn{2}{c}{100--125} &
\multicolumn{2}{c}{125--150} &
\multicolumn{2}{c}{150--175} &
\multicolumn{2}{c}{175$\sim$} \\
\cmidrule(lr){3-4}
\cmidrule(lr){5-6}
\cmidrule(lr){7-8}
\cmidrule(lr){9-10}
\cmidrule(lr){11-12}
\cmidrule(lr){13-14}
\cmidrule(lr){15-16}
& & AUC & TPR@5 & AUC & TPR@5 & AUC & TPR@5 & AUC & TPR@5 & AUC & TPR@5 & AUC & TPR@5 & AUC & TPR@5 \\
\midrule
\multirow{2}{*}{Llama-3.2-3B} & KGW         & 0.997 & 0.984 & 0.997 & 0.992 & 0.996 & 0.990 & 0.996 & 0.990 & 0.996 & 0.990 & 0.996 & 0.990 & 0.997 & 0.990 \\
                              & TripPattern & \textbf{0.998} & \textbf{0.994} & \textbf{0.999} & \textbf{0.996} & \textbf{0.999} & \textbf{0.998} & \textbf{1.000} & \textbf{0.998} & \textbf{1.000} & \textbf{1.000} & \textbf{1.000} & \textbf{1.000} & \textbf{1.000} & \textbf{1.000} \\
\midrule
\multirow{2}{*}{Llama-3.1-8B} & KGW         & \textbf{0.998} & \textbf{0.992} & \textbf{0.999} & \textbf{0.998} & \textbf{0.999} & \textbf{0.998} & 0.999 & \textbf{0.998} & 0.999 & 0.998 & 0.999 & 0.998 & \textbf{1.000} & \textbf{1.000} \\
                              & TripPattern & 0.997 & 0.988 & \textbf{0.999} & 0.996 & \textbf{0.999} & 0.996 & \textbf{1.000} & \textbf{0.998} & \textbf{1.000} & \textbf{1.000} & \textbf{1.000} & \textbf{1.000} & \textbf{1.000} & \textbf{1.000} \\
\midrule
\multirow{2}{*}{Qwen-2.5-3B}  & KGW         & \textbf{0.987} & \textbf{0.974} & \textbf{0.990} & \textbf{0.976} & \textbf{0.992} & \textbf{0.982} & \textbf{0.993} & 0.980 & 0.991 & 0.979 & 0.992 & 0.979 & 0.992 & 0.980 \\
                              & TripPattern & 0.985 & 0.972 & 0.988 & \textbf{0.976} & 0.990 & 0.980 & \textbf{0.993} & \textbf{0.986} & \textbf{0.995} & \textbf{0.992} & \textbf{0.997} & \textbf{0.991} & \textbf{0.997} & \textbf{0.993} \\
\midrule
\multirow{2}{*}{Qwen-2.5-7B}  & KGW         & 0.992 & 0.976 & 0.995 & \textbf{0.992} & 0.997 & 0.990 & \textbf{1.000} & \textbf{0.996} & \textbf{1.000} & \textbf{1.000} & \textbf{1.000} & \textbf{1.000} & \textbf{1.000} & \textbf{1.000} \\
                              & TripPattern & \textbf{0.995} & \textbf{0.984} & \textbf{0.997} & 0.990 & \textbf{0.998} & \textbf{0.996} & 0.998 & \textbf{0.996} & 0.998 & 0.996 & 0.998 & 0.996 & 0.998 & 0.996 \\
\bottomrule
\end{tabular}
}
\end{table*}

\end{document}